\documentclass[11pt]{article}

\usepackage[preprint]{acl}

\usepackage{times}
\usepackage{latexsym}
\usepackage[T1]{fontenc}
\usepackage[utf8]{inputenc}
\usepackage{microtype}
\usepackage{inconsolata}
\usepackage{graphicx}
\usepackage{booktabs}
\usepackage{multirow}
\usepackage{array}
\usepackage{tabularx}
\usepackage{enumitem}
\usepackage{xcolor}
\usepackage{amsmath, amssymb}
\usepackage{hyperref}
\usepackage{url}
\usepackage{tikz}
\usetikzlibrary{arrows.meta,positioning,shapes.geometric,fit}
\usepackage{amssymb}

\definecolor{objectblue}{HTML}{1F77B4}
\definecolor{tracepurple}{HTML}{7B3FB3}
\definecolor{verifygreen}{HTML}{2E8B57}
\definecolor{envteal}{HTML}{168C94}
\definecolor{riskred}{HTML}{B03A2E}

\usepackage{booktabs}
\usepackage{tabularx}
\usepackage{array}
\usepackage[table]{xcolor}
\usepackage{graphicx}

\definecolor{shortcutbg}{RGB}{255,238,235}  
\definecolor{takeawaybg}{RGB}{236,244,255}  
\definecolor{futurebg}{RGB}{232,246,236}    

\newcommand{\shortcuticon}{\raisebox{-0.18em}{\includegraphics[height=1.15em]{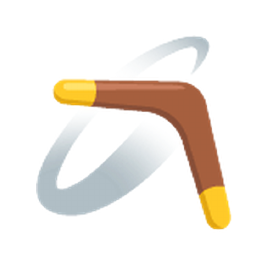}}}
\newcommand{\takeawayicon}{\raisebox{-0.18em}{\includegraphics[height=1.15em]{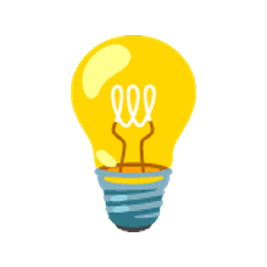}}}
\newcommand{\futureicon}{\raisebox{-0.18em}{\includegraphics[height=1.15em]{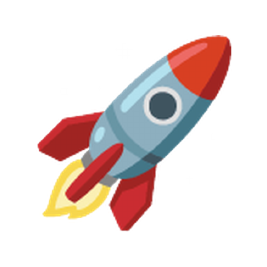}}}

\newcommand{\titleicon}{%
  \raisebox{-0.30em}{%
    \includegraphics[
      height=1.75em,
      trim=0.30in 0.30in 0.30in 0.30in,
      clip
    ]{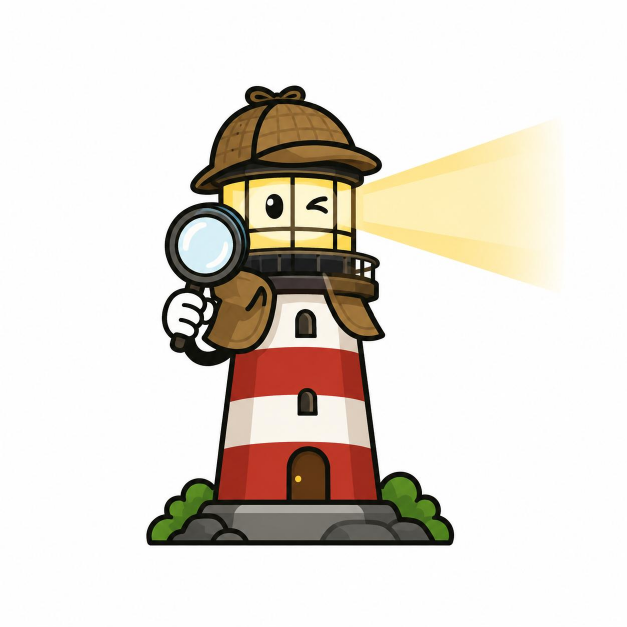}%
  }%
}

\newcommand{\repoicon}{\raisebox{-0.16em}{\includegraphics[height=1.05em]{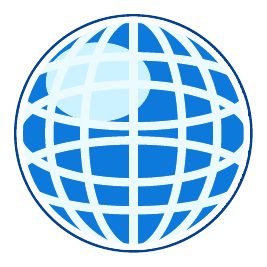}}}

\newcommand{\TakeawayLabel}{\takeawayicon~\textbf{Takeaway}}
\newcommand{\FutureLabel}{\futureicon~\textbf{Future Reporting}}

\newcommand{\SecLink}[1]{\hyperref[#1]{Sec.~\ref*{#1}}}

\newcommand{\secref}[1]{\hyperref[#1]{Sec.~\ref*{#1}}}

\title{\bf
\texorpdfstring{\titleicon\hspace{0.25em}}{}%
A Primer in Post-Training Reasoning Data: What We Know About How It Works}

\author{
 \textbf{Yaoming Li\textsuperscript{1}},
 \textbf{Guangxiang Zhao\textsuperscript{1,3\dag}},
 \textbf{Qilong Shi\textsuperscript{2}},
 \textbf{Lin Sun\textsuperscript{3}},
 \textbf{Xiangzheng Zhang\textsuperscript{3}},
 \textbf{Tong Yang\textsuperscript{1}\thanks{Corresponding authors. \quad \textsuperscript{\dag}Project lead.}},
\\
 \textsuperscript{1}State Key Laboratory of Multimedia Information Processing, \\
 School of Computer Science, Peking University
 \\
 \textsuperscript{2}Tsinghua University
 \textsuperscript{3}Qiyuan Tech
 \\
 {\normalsize \texttt{yibachenggong@gmail.com, zhaoguangxiang@pku.edu.cn, yangtong@pku.edu.cn}}
 \\
 {\normalsize \repoicon\hspace{0.25em}\textbf{Project Repository:} \url{https://github.com/RenBing-Sumeru/Awesome-LLM-Reasoning-Data}}
}

\begin{document}
\maketitle

\begin{abstract}
Post-training has become a primary driver of recent progress in large reasoning models, and reasoning data are often the key variable determining whether this stage succeeds. Work on post-training reasoning data has grown rapidly, yet this literature remains scattered across dataset papers, reinforcement-learning recipes, reward-model studies, benchmarks, and frontier system reports. This paper is the first primer to synthesize over 150 key public studies and system reports on post-training reasoning data. We organize the field around four questions: what data objects exist, what makes them useful, how they are constructed, and how they scale. Together, this organization provides an attribution framework for future reasoning-data releases and post-training recipes.
\end{abstract}

\section{Introduction}
\label{sec:introduction}

The training of large language models typically consists of two stages: pre-training and post-training. With the rise of o1-style test-time scaling paradigms and thinking models, post-training has become increasingly important for further advancing model capabilities \citep{openai_o1_2024,deepseekr12025}. Recent studies suggest that, within the post-training pipeline, the quality and construction of training data often have a greater impact on model performance than other components, such as training algorithms or optimization strategies \citep{openthoughts2025,he2025deepmath103k,xu2025structure}. However, despite the central role of data in post-training, there has not yet been a systematic survey dedicated to post-training data. This paper fills this gap. To the best of our knowledge, it is the first survey focused on post-training data, reviewing 150 key papers in this area and synthesizing practical insights into how post-training data is constructed, curated, and utilized.

\begin{table*}[t]
\centering
\small
\setlength{\tabcolsep}{5pt}
\renewcommand{\arraystretch}{1.08}
\begin{tabularx}{\textwidth}{@{}>{\raggedright\arraybackslash}p{0.22\textwidth}X@{}}
\toprule
\multicolumn{2}{@{}l@{}}{\textbf{Counterintuitive lessons for reasoning-data attribution}} \\
\midrule

\rowcolor{shortcutbg}
\multicolumn{2}{@{}>{\raggedright\arraybackslash}p{\dimexpr\textwidth-2\tabcolsep\relax}@{}}%
{\shortcuticon~\textbf{Trap: ``Long CoT means good reasoning.''} \citep{wei2022cot,turpin2023,lanham2023}} \\
\rowcolor{takeawaybg}
\TakeawayLabel &
Long traces can rationalize, copy teacher style, or hide the actual cause of the answer; quality needs validity and grounding (\secref{subsec:trace-quality}). \\
\rowcolor{futurebg}
\FutureLabel &
Report trace source, process labels, execution context, localization tests, and robustness checks. \\
\midrule

\rowcolor{shortcutbg}
\multicolumn{2}{@{}>{\raggedright\arraybackslash}p{\dimexpr\textwidth-2\tabcolsep\relax}@{}}%
{\shortcuticon~\textbf{Trap: ``Harder data is better data.''} \citep{he2025deepmath103k,bigmath2025,dapo2025}} \\
\rowcolor{takeawaybg}
\TakeawayLabel &
Difficulty is base-relative: a problem can be unreachable for one base, useful for another, and saturated for a third (\secref{subsec:difficulty-base-relative}). \\
\rowcolor{futurebg}
\FutureLabel &
Report the base model, sampling protocol, verifier, rollout count, temperature, pass-rate band, and estimate date. \\
\midrule

\rowcolor{shortcutbg}
\multicolumn{2}{@{}>{\raggedright\arraybackslash}p{\dimexpr\textwidth-2\tabcolsep\relax}@{}}%
{\shortcuticon~\textbf{Trap: ``More data means better coverage.''} \citep{openthoughts2025,xu2025structure,shumailov2024collapse}} \\
\rowcolor{takeawaybg}
\TakeawayLabel &
Coverage is a recipe, not a count; source mixture, filters, teachers, generators, leakage, and lineage decide what is inherited (\secref{subsec:coverage-lineage}). \\
\rowcolor{futurebg}
\FutureLabel &
Report source mixture, generator, teacher, filtering rule, split, decontamination status, and lineage risks. \\
\midrule

\rowcolor{shortcutbg}
\multicolumn{2}{@{}>{\raggedright\arraybackslash}p{\dimexpr\textwidth-2\tabcolsep\relax}@{}}%
{\shortcuticon~\textbf{Trap: ``A clean successful transcript is ideal.''} \citep{toolllm2023,swegym2024,osworld2024}} \\
\rowcolor{takeawaybg}
\TakeawayLabel &
For agents, cleaned success traces may erase failures, retries, recoveries, and state diffs---the evidence for credit assignment (\secref{subsec:search-substrate}). \\
\rowcolor{futurebg}
\FutureLabel &
Release replayable episodes with states, actions, observations, failures, retries, terminal predicates, and scaffold metadata. \\
\midrule

\rowcolor{shortcutbg}
\multicolumn{2}{@{}>{\raggedright\arraybackslash}p{\dimexpr\textwidth-2\tabcolsep\relax}@{}}%
{\shortcuticon~\textbf{Trap: ``Self-play removes curation.''} \citep{zelikman2022star,zhao2025absolutezero,zuo2025ttrl}} \\
\rowcolor{takeawaybg}
\TakeawayLabel &
Self-play relocates curation to the anchor---answer, interpreter, verifier, majority vote, archive, or role split (\secref{subsec:self-play}). \\
\rowcolor{futurebg}
\FutureLabel &
Report the anchor, selection rule, verifier, admissibility rule, replay policy, and failure modes. \\
\midrule

\rowcolor{shortcutbg}
\multicolumn{2}{@{}>{\raggedright\arraybackslash}p{\dimexpr\textwidth-2\tabcolsep\relax}@{}}%
{\shortcuticon~\textbf{Trap: ``The optimizer explains the gain.''} \citep{shao2024grpo,deepseekr12025,dapo2025}} \\
\rowcolor{takeawaybg}
\TakeawayLabel &
Optimizer visibility is not causal isolation; the same RL scaffold can hide different prompt pools, trace writers, substrates, and verifiers (\secref{sec:how-built}). \\
\rowcolor{futurebg}
\FutureLabel &
Disclose prompt source, trace author, search substrate, reward channel, scaffold, and budget. \\
\midrule

\rowcolor{shortcutbg}
\multicolumn{2}{@{}>{\raggedright\arraybackslash}p{\dimexpr\textwidth-2\tabcolsep\relax}@{}}%
{\shortcuticon~\textbf{Trap: ``Scaling means the model got better.''} \citep{khatri2025scalerl,tan2025scalingrl,hochlehnert2025sober}} \\
\rowcolor{takeawaybg}
\TakeawayLabel &
A higher score may move the ceiling, improve efficiency, or change the measurement surface; these are different claims (\secref{subsec:asymp-eff}). \\
\rowcolor{futurebg}
\FutureLabel &
Report unique data, reuse rate, training compute, inference budget, search topology, verifier refresh, and evaluation protocol. \\

\bottomrule
\end{tabularx}
\caption{\textbf{Counterintuitive lessons for reasoning-data attribution.}
The table distills recurring traps in post-training reasoning-data work, why they fail, links each takeaway to the section where it is developed, and lists the future reporting fields needed to make gains attributable.}
\label{tab:counterintuitive-lessons}
\vspace{-0.45cm}
\end{table*}

\begin{figure*}[t]
    \centering
    \includegraphics[width=\textwidth]{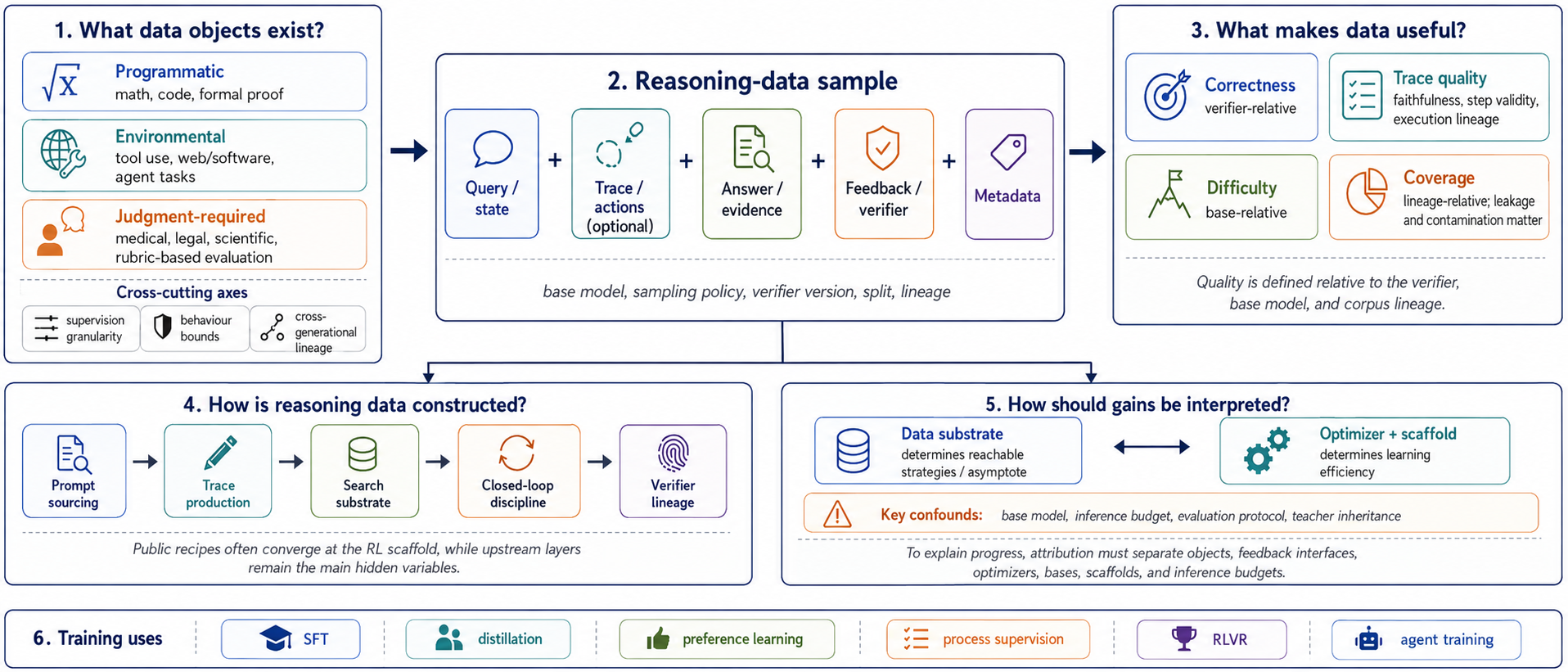}
    \caption{\textbf{Beyond prompt--response pairs.}
    A reasoning-data item packages a problem or state, model behavior, judging feedback, and attribution metadata. The map previews four questions: objects, usefulness, construction, and gain attribution.}
    \label{fig:feedback-interface}
    \vspace{-0.45cm}
\end{figure*}



Post-training reasoning data is therefore not just a collection of datasets, but a set of evidence-bearing records for explaining why a model improves. As summarized in Table~\ref{tab:counterintuitive-lessons}, many intuitive assumptions about reasoning data are misleading: long CoT traces do not necessarily imply faithful reasoning \citep{wei2022cot,turpin2023,lanham2023}; harder data is useful only relative to a base model and verifier \citep{he2025deepmath103k,bigmath2025,dapo2025}; larger corpora may still provide poor coverage when source mixture, leakage, or lineage are uncontrolled \citep{openthoughts2025,xu2025structure,shumailov2024collapse}; and, in agentic settings, overly cleaned success trajectories can remove precisely the failures, retries, and state changes needed for credit assignment \citep{toolllm2023,swegym2024,osworld2024}. These pitfalls reveal a missing common language for connecting data objects, feedback signals, construction recipes, optimizers, base models, scaffolds, and budgets. We organize the primer around four questions---what reasoning-data objects exist, what makes them useful, how they are constructed, and how their gains scale. As Figure~\ref{fig:feedback-interface} shows, this framing moves beyond prompt--response pairs and treats each item as a feedback-bearing record linking task context, model behavior, judging feedback, and attribution metadata. The goal is not just to enumerate datasets, but to help readers inspect future releases and attribute reported gains.

\section{What Data Do We Need? A Verifier-Anchored Taxonomy}
\label{sec:what-data-need}

\begin{figure}[t]
    \centering
    \includegraphics[width=\columnwidth]{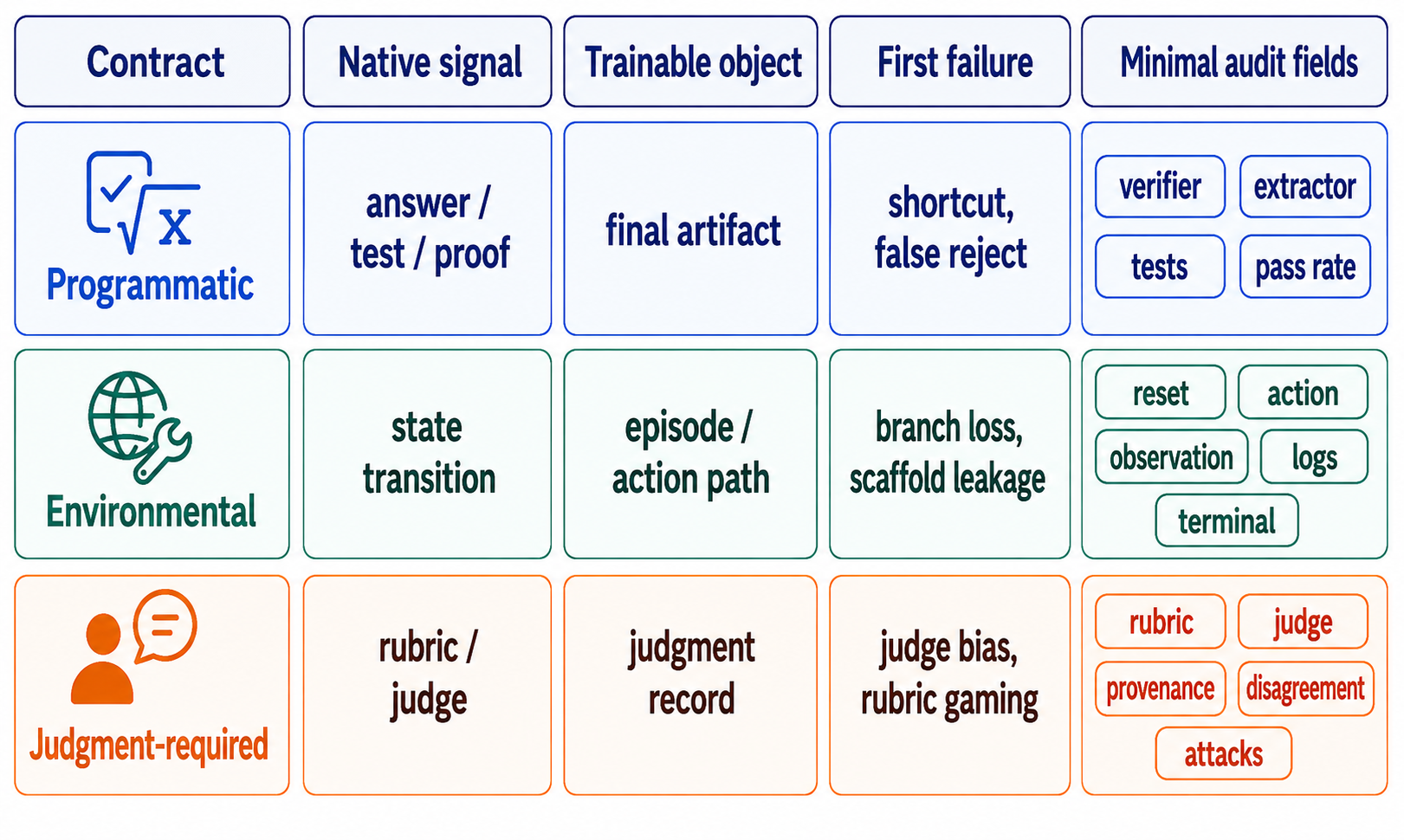}
    \caption{\textbf{Verifier-anchored taxonomy.}
    Verification contracts, rather than domains, define the native signal, trainable object, failure surface, and required audit fields of a reasoning sample.}
    \label{fig:verifier-taxonomy}
    \vspace{-0.4cm}
\end{figure}

Reasoning corpora are usually named by domain, but the learning signal is defined by what can be checked. We therefore sort data by \emph{verification contract}: where feedback enters, what becomes trainable, and which fields make the signal auditable. The unit is a \emph{verifier-bearing sample}, not a domain-labelled prompt--response pair \citep{lightman2023verify,mathshepherd2023}. Figure~\ref{fig:verifier-taxonomy} gives the taxonomy as an attribution lens: it specifies what supervision a sample exposes, not whether the trained model later uses that supervision at inference time.

\subsection{Three Verification Contracts}
\label{subsec:verification-contracts}

\paragraph{Programmatic verification.}
Programmatic contracts make artifacts checkable: answers can be normalized, code executed, and proof states verified, enabling scalable math, code, and Lean corpora \citep{openmathreasoning2025,ahmad2025opencodereasoning2,kodcode2025,deepseekproverv2_2025}. Signals are local: verifiers may certify a final artifact without certifying robustness, non-spuriousness, or movement beyond the base policy \citep{spuriousrewards2025,yue2025rlreasoning,helff2026gamingverifiers}. Since small curated sets can elicit latent skills \citep{s12025,limo2025}, releases should report checked object, extractor, filter, and base-conditioned difficulty.

\paragraph{Environmental verification.}
Environmental contracts make interaction trainable. Tool, web, app, OS, and repository tasks expose goals, actions, observations, files, state transitions, and terminal predicates \citep{toolllm2023,mind2web2023,weblinx2024,visualwebarena2024,appworld2024,taubench2024,swebenchverified2024,r2egym2025}. Their strength is state-based success; their failure is transcript compression. Clean SFT traces often remove failed actions, retries, recoveries, and sibling rollouts---the branches where credit assignment is visible. Such data should be replayable episodes, not only successful transcripts.

\paragraph{Judgment-required verification.}
When no deterministic verifier exists, the reusable unit is an auditable judgment record. Medical, factuality, safety, and rubric-reward datasets attach criteria, evidence, risk, provenance, judge traces, or disagreement fields \citep{healthbench2025,bao2025faithbench,ghosh2025aegis2,sreedhar2025safetyreasoning,anugraha2025r3}. This expands scope, but makes the judge part of the data object: hidden prompts, rubrics, calibration sets, or judge identities can turn gains into judge-specific preferences \citep{zhao2025masterrm}. Releases should ship the judgment contract itself.

\subsection{Three Cross-Cutting Axes}
\label{subsec:cross-cutting-axes}

\begin{figure}[t]
    \centering
    \includegraphics[width=\columnwidth]{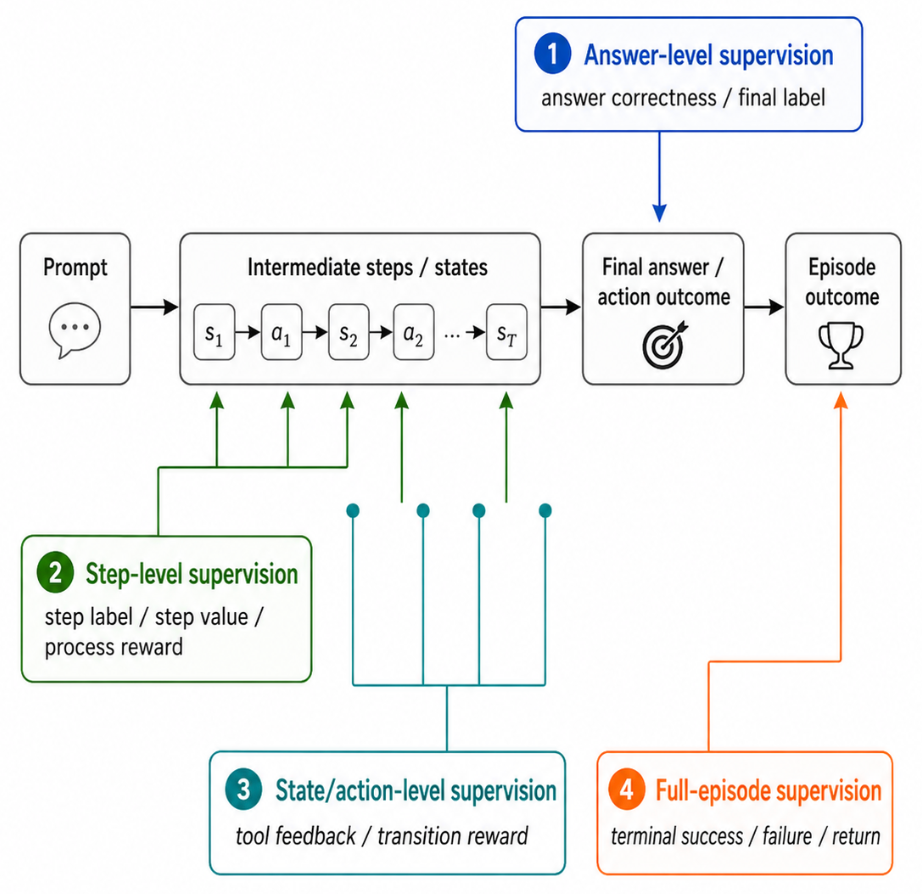}
    \caption{\textbf{Where supervision enters the trajectory.}
    The same task becomes different post-training data when feedback targets the final answer, an intermediate step, a state--action transition, or the full episode.}
    \label{fig:supervision-entry}
    \vspace{-0.5cm}
\end{figure}

The contract says what verifies a sample; the next question is where that feedback enters the trajectory. Figure~\ref{fig:supervision-entry} shows four entry points: feedback on the final answer, an intermediate reasoning step, a state--action transition, or the full episode. Moving the feedback point changes the trainable object even when the underlying task is the same.

\paragraph{Supervision granularity.}
Outcome labels check endpoints; step labels localize valid reasoning; continuation scores estimate stability; formal and tool traces make operations reproducible \citep{lightman2023verify,mathshepherd2023,omegaprm2024,openhands2025,kim2026selfdistillationreasoning}. This separates \emph{capability} from \emph{mechanism}: accuracy may show that a task is solved, while the teachable signal may live in steps, rollouts, or execution.

\paragraph{Behaviour bounds.}
Behaviour-bounding data define when to answer, refuse, abstain, provide safe completion, or expose a trace \citep{orbench2025,kirichenko2025abstentionbench,sreedhar2025safetyreasoning,jiang2025safechain}. The caution is that abstention, verbosity, or exposed chains need not imply reliability \citep{turpin2023,lanham2023}; releases should separate intent, risk, allowed content, response type, trace visibility, and mismatch.

\paragraph{Cross-generational lineage.}
Synthetic flywheels make data cheap, but what propagates may be trace style, decoding policy, filter, or teacher preference rather than the question distribution \citep{openthoughts2025,zhao2026decouplekl}. Scale can coexist with narrowing, hidden trait transfer, leakage, or search contamination \citep{shumailov2024collapse,green2025leakythoughts,fang2025lastingbench,han2025searchcontamination}. Lineage should therefore be sample-level metadata.

Together, these axes explain why contract labels alone do not determine how the signal behaves. A sample can be programmatically verified but answer-only, environmentally verified but scaffold-inherited, or judgment-required but shaped by teacher lineage.

The taxonomy yields four downstream gaps: value is base-relative; environmental signal may live in branches and failures; verifier errors must be attributable to the model, verifier, scaffold, or schema \citep{helff2026gamingverifiers,plesner2026imperfectverifier}; and leakage must be audited beyond answer strings. They motivate the next questions: what makes a sample useful, how it is built, and how scaling claims should be attributed.

\section{What Makes Reasoning Data Good?}
\label{sec:good-data}

\begin{figure}[t]
    \centering
    \includegraphics[width=\columnwidth]{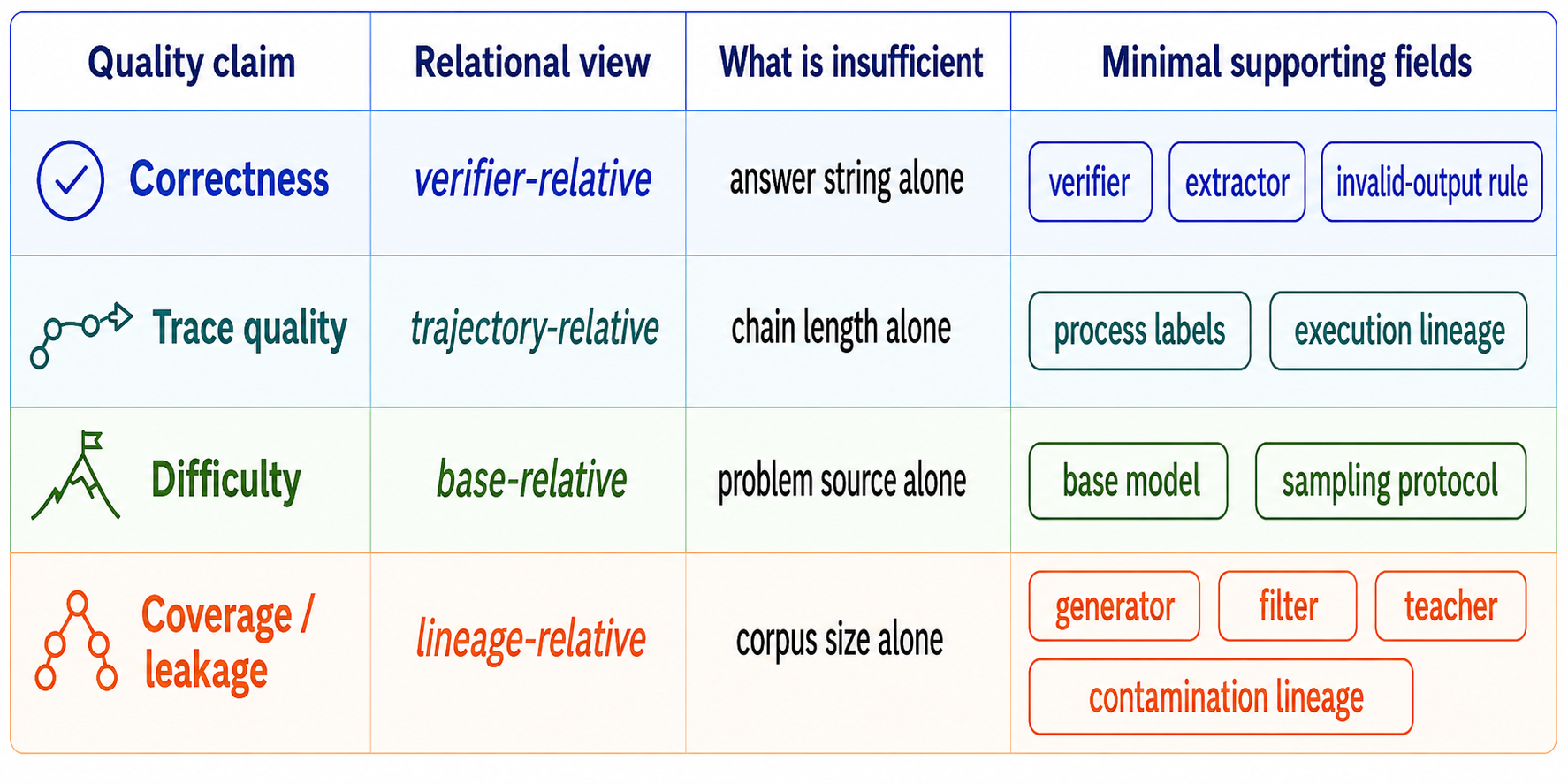}
    \caption{\textbf{Quality support matrix.}
    No single field licenses all quality claims: correctness, difficulty, trace quality, and coverage each require different verifier, base, trajectory, and lineage evidence.}
    \label{fig:quality-diagnostic-map}
    \vspace{-0.45cm}
\end{figure}

Section~\ref{sec:what-data-need} treated reasoning data as verifier-bearing samples; the quality question is therefore which attached fields license a claim about the sample. Recent releases expose different parts of this claim: DeepMath-103K and DAPO make extraction and rule verification operational \citep{he2025deepmath103k,dapo2025}; PRM800K and Math-Shepherd move supervision into steps and rollouts \citep{lightman2023verify,mathshepherd2023}; Skywork-OR1 makes difficulty model-aware \citep{he2025skyworkor1}; and OpenThoughts makes data construction ablatable \citep{openthoughts2025}. Figure~\ref{fig:quality-diagnostic-map} summarizes the view: correctness is verifier-relative, difficulty is base-relative, trace quality is trajectory-relative, and coverage is lineage-relative.

\subsection{Correctness Is Verifier-Relative}
\label{subsec:correctness-verifier-relative}

Correctness is a versioned verifier contract, not an answer string. DeepMath-103K and DAPO show that extraction, normalization, rule checking, invalid-output handling, and pass-rate filtering determine which rollouts become trainable \citep{he2025deepmath103k,dapo2025}. Yet verifiers are also failure surfaces: rule verifiers can reject equivalent answers, and model judges can be steered by superficial cues or verifier-gaming policies \citep{tinyv2025,huang2025verifierrobustness,zhao2025masterrm,helff2026gamingverifiers}. Correctness claims should therefore report the extractor, verifier version, filtering rule, and known failure modes.

\subsection{Good Traces Are Not Just Long Traces}
\label{subsec:trace-quality}

\begin{figure}[t]
    \centering
    \includegraphics[width=\columnwidth]{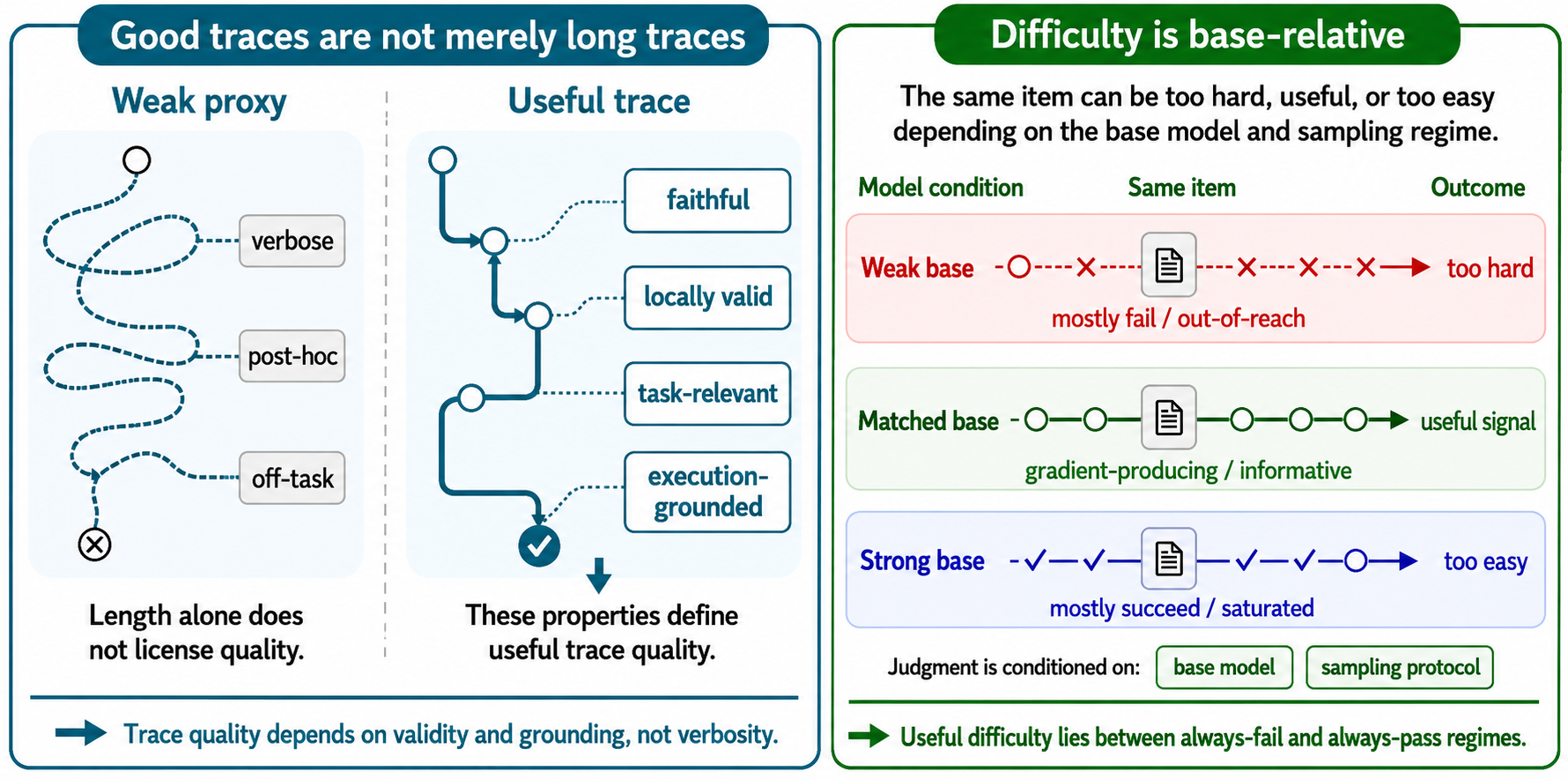}
    \caption{\textbf{Two common quality traps: long trace $\neq$ good trace, hard item $\neq$ useful item.}
    Trace quality depends on validity and grounding, while useful difficulty lies between always-fail and always-pass regimes for a given base and sampling protocol.}
    \label{fig:good-trace-not-long}
    \vspace{-0.45cm}
\end{figure}

A visible trace is suggestive evidence, not proof of mechanism. Rationales may omit the factors that drove the answer or rationalize decisions shaped by biased features \citep{lanham2023,turpin2023}. Figure~\ref{fig:good-trace-not-long} makes the distinction explicit: useful traces are auditable, locally valid, task-relevant, and grounded in labels or execution context.

Process supervision makes intermediate validity observable, but it does not remove the need for audit: PRM800K, Math-Shepherd, and OmegaPRM differ in whether the signal comes from human labels, rollout values, or refined process estimates \citep{lightman2023verify,mathshepherd2023,omegaprm2024}. ProcessBench and PRMBench further show that PRMs must be tested for localization, soundness, sensitivity, and robustness \citep{processbench2025,prmbench2025}. For tool traces, sandbox, timeout, package version, output, and execution lineage are part of the sample \citep{openmathreasoning2025}.

\subsection{Difficulty Is Base-Relative}
\label{subsec:difficulty-base-relative}

Difficulty is an item--model--sampling relation. Figure~\ref{fig:good-trace-not-long} illustrates the corresponding trap: a hard item is not necessarily a useful item, because it may be unreachable for one base, gradient-producing for another, and saturated for a third.

Skywork-OR1 estimates difficulty against model variants, while DAPO implements the same idea by removing prompt groups with accuracy $0$ or $1$ \citep{he2025skyworkor1,dapo2025}. Thus ``medium difficulty'' hides the base model, prompt format, rollout count, temperature, and verifier. Because scores can shift with seeds, rollouts, and token budgets, and because RLVR may reweight base-supported trajectories rather than expand the frontier, difficulty should be released as a dated base-conditional estimate \citep{hochlehnert2025sober,yue2025rlreasoning}.

\subsection{Coverage, Leakage, and Lineage}
\label{subsec:coverage-lineage}

Coverage is a recipe, not a count. OpenThoughts ablates source, mixture, filter, generator, and teacher choices \citep{openthoughts2025}; Structure Trumps Size shows that data structure can matter more than conventional cleanliness or scale \citep{xu2025structure}. Synthetic data make this recipe inherit across generations: corpora can narrow recursively, transmit hidden teacher traits, leak private fields through traces, or contaminate benchmarks through search channels \citep{shumailov2024collapse,cloud2025subliminal,green2025leakythoughts,han2025searchcontamination}. A useful release should therefore report generator, filter, split, teacher, verifier of record, decontamination status, and known inheritance risks.

\section{How Is Post-Training Reasoning Data Built?}
\label{sec:how-built}

Public reports have made the RL scaffold increasingly visible: long-CoT SFT, distillation, RLVR, reward modelling, and GRPO-style optimization recur across DeepSeek-R1, Kimi K1.5, Qwen3, Magistral, Phi-4-reasoning, and Llama-Nemotron \citep{deepseekr12025,kimi2025,yang2025qwen3,mistral2025magistral,abdin2025phi4reasoning,bercovich2025llamanemotron,shao2024grpo}. But optimizer visibility is not causal isolation. DAPO changes filtering and loss geometry, MiniMax-M1 changes importance-weight control, and Qwen3-Coder-style systems move signal into long-horizon tool interaction \citep{dapo2025,minimax2025m1,qwen3coder2025}. Table~\ref{tab:construction-fields} therefore treats reasoning-data construction as an attribution ledger: each upstream layer exposes a visible field, hides a different confound, and determines which metadata are needed to interpret the final gain.

\begin{table}[t]
\centering
\scriptsize
\setlength{\tabcolsep}{2.8pt}
\renewcommand{\arraystretch}{1.10}
\begin{tabularx}{\columnwidth}{@{}
>{\raggedright\arraybackslash}p{0.27\columnwidth}
>{\raggedright\arraybackslash}X
ccccc@{}}
\toprule
\textbf{Build layer} 
& \textbf{Attribution handle} 
& \textbf{Q} & \textbf{T} & \textbf{E} & \textbf{V} & \textbf{L} \\
\midrule
Prompt sourcing
& problem support / pass-rate band
& $\checkmark$ & -- & -- & $\triangle$ & $\triangle$ \\

Trace writing
& inherited reasoning style
& $\triangle$ & $\checkmark$ & -- & $\triangle$ & $\checkmark$ \\

Search substrate
& exploration and replayability
& $\triangle$ & $\triangle$ & $\checkmark$ & $\checkmark$ & $\triangle$ \\

Self-play anchor
& where curation re-enters
& $\checkmark$ & $\triangle$ & $\triangle$ & $\checkmark$ & $\checkmark$ \\

Reward/verifier
& what counts as success
& $\triangle$ & $\triangle$ & $\triangle$ & $\checkmark$ & $\checkmark$ \\

Frontier pipeline
& where reports appear to converge
& $\checkmark$ & $\checkmark$ & $\checkmark$ & $\checkmark$ & $\checkmark$ \\
\bottomrule
\end{tabularx}
\caption{\textbf{Construction as an attribution ledger.}
Each build layer exposes a different handle for attributing the final gain. 
$\checkmark$ marks the primary trainable field, $\triangle$ an audit field, and -- a usually absent or non-primary field. 
Q = prompt/source; T = trace or teacher; E = environment/substrate; V = verifier/reward; L = lineage.}
\label{tab:construction-fields}
\vspace{-0.45cm}
\end{table}

\subsection{Where Do Reasoning Prompts Come From?}
\label{subsec:prompts}

Prompt sourcing is where a recipe first becomes data-specific. OpenThoughts makes source, mixture, filtering, answer generation, and teacher choice explicit ablation variables \citep{openthoughts2025}; math and code corpora turn olympiad pools, web-mined problems, executable tests, and model-conditioned solve rates into prompt-side fields \citep{bigmath2025,openmathreasoning2025,he2025deepmath103k,naturalreasoning2025,ahmad2025opencodereasoning2}. Small-set results such as LIMO and s1 provide the counterpoint: few demonstrations can elicit strong behaviour only when the skill already lies within the base policy's support \citep{limo2025,s12025,yue2025rlreasoning,wu2025invisibleleash}. Prompt releases should therefore report source, filter, verifier, per-base pass rate, and estimate date, not only domain and count.

\subsection{Who Writes the Trace?}
\label{subsec:trace-writer}

Trace construction fixes the author, process field, and grounding of the trajectory. Teacher traces can transfer decomposition style, uncertainty expression, formatting, tool use, and stopping behaviour, not merely answers \citep{openthoughts2025,ji2025amthinkingv1,abdin2025phi4reasoning,kim2026selfdistillationreasoning}. Process-supervised variants make hidden structure more explicit: PRM800K labels steps, Math-Shepherd estimates rollout value, OmegaPRM searches for first errors, PRIME derives implicit process reward, and PROF filters incoherent chains \citep{lightman2023verify,mathshepherd2023,omegaprm2024,prime2025,ye2025prof}. The dispute is therefore not whether traces matter, but whether outcome-only RL, explicit PRMs, or trainable verifiers expose the relevant process field \citep{sullivan2025grpoprm,deepseekmathv2_2025}.

\subsection{Where Do Search-Time Gains Live?}
\label{subsec:search-substrate}

Agentic reasoning moves search from decoding into an executable substrate. Repository, web, app, OS, and tool benchmarks define samples through states, actions, observations, files, tests, and terminal predicates \citep{swegym2024,r2egym2025,appworld2024,taubench2024,osworld2024,openhands2025}; theorem-proving systems collapse action and verification into Lean-state transitions \citep{deepseekproverv2_2025,lin2025goedelproverv2}. The lesson is not that every task needs an environment, but that environment data should be replayable. Successful transcripts alone erase failed actions, recoveries, retries, state diffs, and hidden predicates---the branches where credit assignment is visible.

\subsection{Can Self-Play Eliminate Curation?}
\label{subsec:self-play}

Self-play does not eliminate curation; it moves curation into the anchor that turns generated behaviour into trainable feedback. STaR anchors rollouts to external answers, R-Zero separates Challenger and Solver roles, Absolute Zero uses the Python interpreter, TTRL turns test-time majority into reward, and multi-agent variants distribute these functions across roles \citep{zelikman2022star,huang2025rzero,zhao2025absolutezero,zuo2025ttrl,multiagentevolve2025}. These systems differ less in agent count than in what makes a trajectory admissible: answer availability, executable feedback, majority selection, role-mediated challenge, or archive-based evaluation.

The optimistic reading is autonomous curriculum growth. The caution is that anchors also define support bounds and label reliability: on-policy RLVR may conservatively reweight trajectories already supported by the base policy, and pseudo-label accuracy can degrade as generated tasks harden \citep{wu2025invisibleleash,huang2025rzero}. AlphaEvolve illustrates the counter-pressure: discovery is auditable precisely because the evaluator and evolutionary archive remain explicit anchors \citep{novikov2025alphaevolve}. Self-play releases should therefore report the anchor, selection rule, verifier, replay policy, and failure modes, not only generated tasks or final success rates.

\subsection{What Does the Policy Train Against?}
\label{subsec:verifier}

\begin{figure}[t]
    \centering
    \includegraphics[width=0.96\columnwidth]{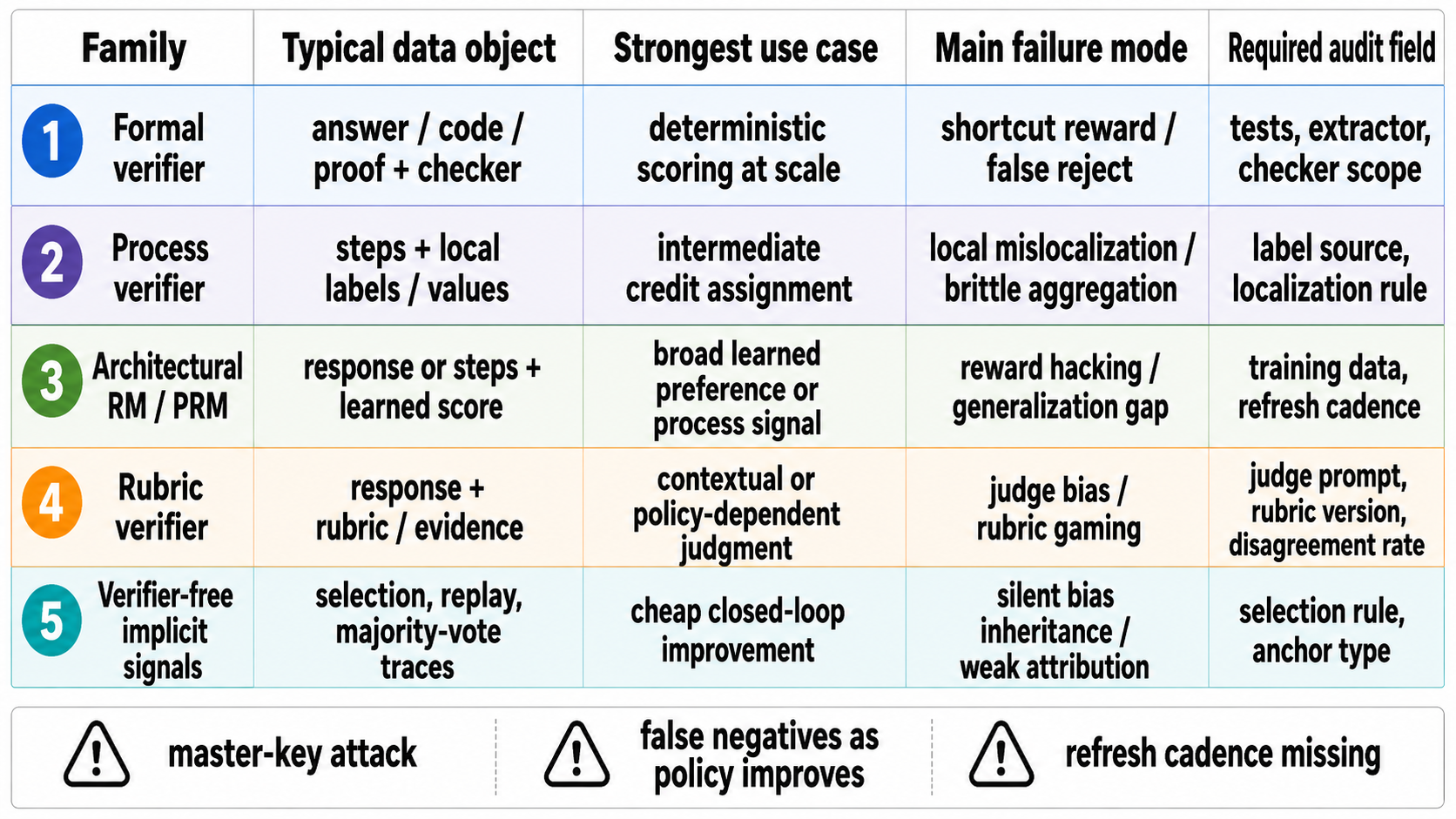}
    \caption{\textbf{Verifier families and failure surfaces.}
    A reward channel is itself a data object: formal checkers, process verifiers, learned reward models, rubric judges, and implicit selectors expose different use cases, failure modes, and audit fields.}
    \label{fig:verifier-failures}
    \vspace{-0.45cm}
\end{figure}

The reward channel is also a data object: a ``reward'' may be a formal checker, process verifier, learned reward model, rubric judge, or closed-loop selection rule, as summarized in Figure~\ref{fig:verifier-failures}.

Master-key attacks, spurious rewards, GSM-Symbolic perturbations, and verifier-gaming tests all show that reward signals can be broad yet brittle \citep{zhao2025masterrm,shao2025spuriousrewards,gsmsymbolic2024,helff2026gamingverifiers}. Rubric systems extend the same problem to domains without symbolic checking: HealthBench, RaR, OnlineRubrics, Omni-RM, AutoRubric, RubricArm, and PoP make criteria, elicitation, modality, and judge-policy interaction part of the reward-data pipeline \citep{healthbench2025,rar2025,rezaei2025onlinerubrics,kong2026omnirrm,rao2026autorubric,xu2026rubricarm,huang2026pop}. CoVerRL and DeepSeekMath-V2 suggest co-evolving generators and verifiers, but co-evolution raises the need for versioning rather than removing it \citep{pan2026coverrl,deepseekmathv2_2025}.

\subsection{What Does a Frontier Pipeline Look Like?}
\label{subsec:frontier-pipeline}

A frontier pipeline is an orchestration of the fields above. Distill-then-RL systems buy cold-start speed through teacher traces; small-warmup multi-stage RL systems let RL reshape the policy; pure-RL lines reduce teacher inheritance but increase dependence on verifier and prompt support \citep{abdin2025phi4reasoning,bercovich2025llamanemotron,deepseekr12025,yang2025qwen3,mistral2025magistral,hu2025openreasonerzero}. Agent-native systems add tool calls, execution feedback, long contexts, and software environments to the post-training distribution itself \citep{moonshot2025kimik2,minimax2025m1,qwen3coder2025}. The optimizer is therefore the wrong final unit of comparison: a reported gain is interpretable only after the prompt support, trace teacher, substrate, anchor, verifier, scaffold, and inference budget are declared.

\section{Scaling: Asymptotes on the Data Substrate, Efficiency on the Optimiser}
\label{sec:scaling}

Recent scaling sweeps make reasoning post-training look less like a single law than a ledger of what changed. \citet{khatri2025scalerl} separate asymptotic performance from compute efficiency in large-scale RL runs, while \citet{tan2025scalingrl} show that model size, compute, and data reuse interact across the Qwen2.5 dense series. We use their equations as a reading device rather than a universal law:
\begin{equation}
\label{eq:scaling-ledger}
\begin{gathered}
R(C) = R_0 + \frac{A-R_0}{1+(C_{\mathrm{mid}}/C)^B} \\[0.35em]
\log L(N,C) = E(N) - k(N)\log C
\end{gathered}
\end{equation}
Here $A$ is the reachable ceiling, while $B$ or $k(N)$ captures approach efficiency. A benchmark gain is therefore not self-explanatory: it may move the ceiling, improve approach speed, or change the measurement surface.

\begin{figure}[t]
    \centering
    \includegraphics[width=0.96\columnwidth]{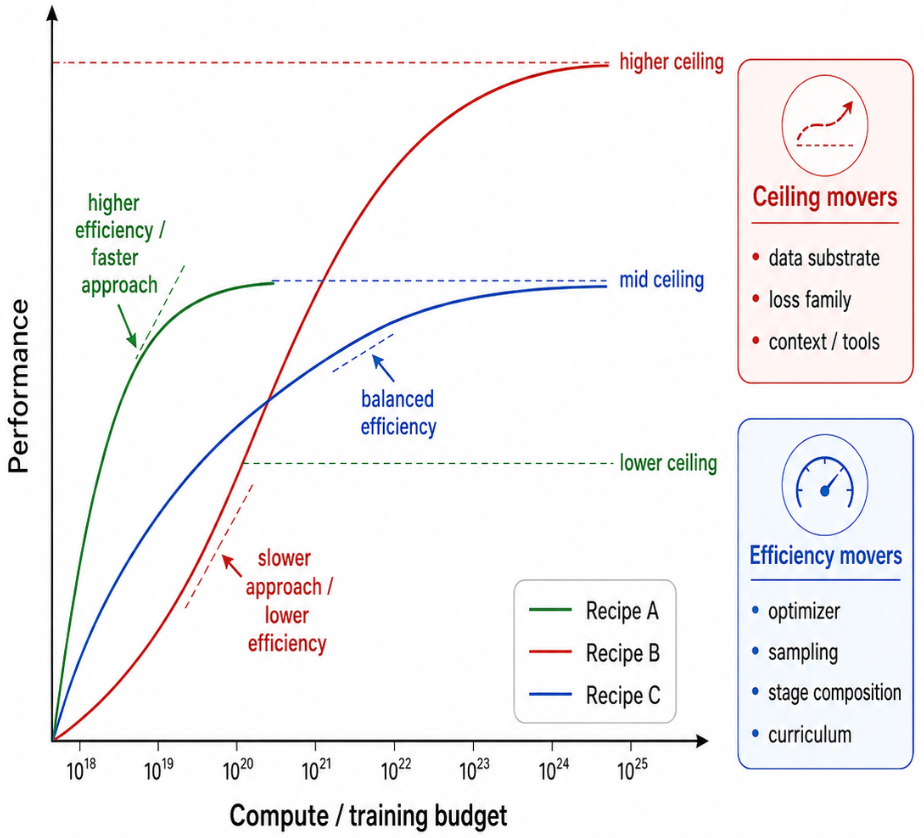}
    \caption{\textbf{Asymptote--efficiency decomposition.}
    A scaling result can change what the data substrate makes reachable, how efficiently training approaches that frontier, or both.}
    \label{fig:asymptote-efficiency}
    \vspace{-0.35cm}
\end{figure}

\begin{table}[t]
\centering
\small
\setlength{\tabcolsep}{2.6pt}
\renewcommand{\arraystretch}{1.08}
\begin{tabularx}{\columnwidth}{@{}
>{\raggedright\arraybackslash}p{0.34\columnwidth}
cc
>{\raggedright\arraybackslash}X@{}}
\toprule
\textbf{Scaling knob} 
& \textbf{$A$} 
& \textbf{$B/k$} 
& \textbf{Release fields} \\
\midrule
Prompt/filter pool     
& $\checkmark$ & $\triangle$  
& source, filter, pass-rate band \\

Loss / clipping / entropy 
& $\triangle$ & $\checkmark$ 
& objective, KL, batch, weighting \\

Reuse / epochs         
& $\triangle$ & $\checkmark$ 
& $D_{\mathrm{unique}}$, $\tau$, replay rule \\

Search / TTT topology  
& $\triangle$ & $\checkmark$ 
& budget rule, selector, verifier, metric \\

Environment substrate  
& $\checkmark$ & $\triangle$  
& state, actions, replay, terminal check \\

Verifier refresh       
& $\checkmark$ & $\checkmark$ 
& verifier card, attacks, refresh log \\
\bottomrule
\end{tabularx}
\caption{\textbf{Scaling attribution ledger.}
$\checkmark$ marks the dimension a knob most directly affects, and $\triangle$ a conditioning effect. 
$A$ = reachable ceiling; $B/k$ = approach efficiency. 
Search topology, inference budget, and evaluation protocol may change the measured capability, so they should be distinguished from training compute.}
\label{tab:scaling-ledger}
\vspace{-0.45cm}
\end{table}

\subsection{Asymptotes and Efficiency}
\label{subsec:asymp-eff}

The useful commonality between the Khatri and Tan laws is the separation in Figure~\ref{fig:asymptote-efficiency}: some choices change the reachable frontier, while others change the path toward it \citep{khatri2025scalerl,tan2025scalingrl}. Table~\ref{tab:scaling-ledger} gives the audit reading. Data substrate, verifier quality, support coverage, context, architecture, and search topology can move the ceiling; loss design, sampling, rollout budget, curriculum, precision, and warm-start distillation more directly alter efficiency.

This distinction also organizes the RLVR debate. \citet{yue2025rlreasoning} and \citet{wu2025invisibleleash} read current RLVR as sharpening trajectories already accessible to the base policy. The counter-literature relaxes different closures: ProRL extends horizon length \citep{prorl2025}; RL-PLUS injects external rollouts \citep{rlplus2025}; CoT-Pass@K changes the success metric \citep{wen2025cotpassk}; and PASS@$(k,T)$ adds interaction depth \citep{zhai2026passkt}. Thus ``RL expands capability'' is not a scalar claim; it specifies which closure was broken.

\subsection{Data Uniqueness and Stage Composition}
\label{subsec:data-uniqueness}

\begin{figure}[t]
    \centering
    \includegraphics[width=0.98\columnwidth]{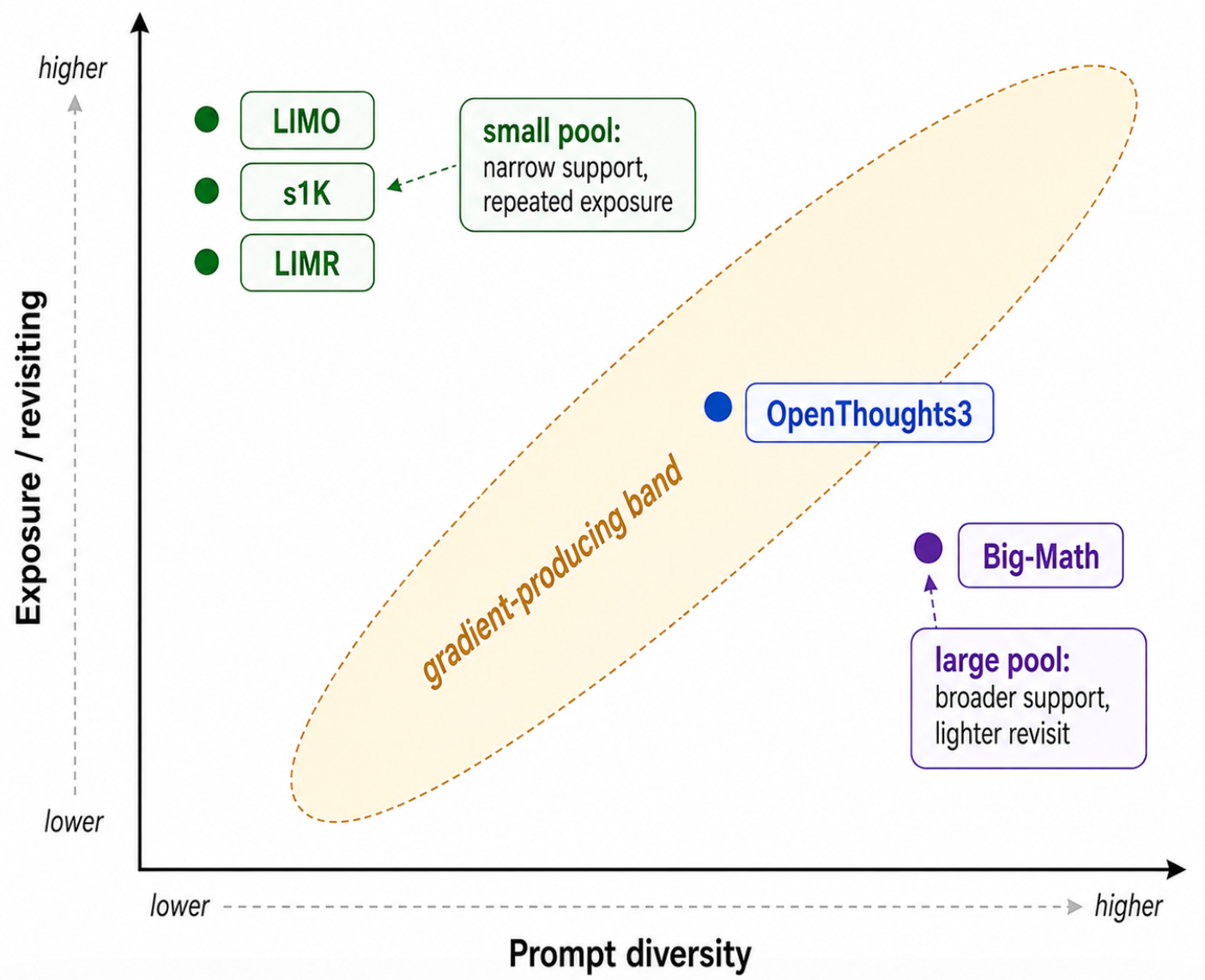}
    \caption{\textbf{Small-pool versus large-pool coverage.}
    Small curated pools can be effective when they repeatedly sample the active band of a capable base; larger pools matter when useful gradients lie in the tail or must cover multiple bases, verifiers, or domains.}
    \label{fig:pool-coverage-band}
    \vspace{-0.5cm}
\end{figure}

Tan et al.'s decomposition $D_{\mathrm{total}}=D_{\mathrm{unique}}\times\tau$ makes reuse part of the scaling surface rather than a preprocessing detail \citep{tan2025scalingrl}. Figure~\ref{fig:pool-coverage-band} reframes the small-pool versus large-pool debate: what matters is whether examples fall inside the gradient-producing band for a given base model.

Small-pool releases such as LIMO and s1 show hundreds or thousands of examples can elicit strong behaviour when the base already supports the skill \citep{limo2025,yue2025rlreasoning,wu2025invisibleleash}. OpenThoughts and Big-Math expose the opposite regime, where source mixture, filtering, teacher choice, and base-conditional solve rates determine coverage \citep{openthoughts2025,bigmath2025}. Cross-base studies sharpen the point: cognitive behaviours and even spurious rewards are base-conditional rather than generic \citep{gandhi2025cognitive,shao2025spuriousrewards}. Distillation and stage composition decide what enters RL before scaling is measured, since teacher traces, grounded thoughts, mid-training, SFT, and RL prefer different quality--diversity trade-offs \citep{busbridge2025distillationscaling,tu2025midtrainingsurvey,mo2025midtrainingsurvey}. Scaling is therefore parameter count under a base prior, uniqueness budget, teacher lineage, and stage order.

\subsection{Inference-Time Compute and Trace Topology}
\label{subsec:inference-time}

Inference-time scaling changes the observable, not merely the decoding setting. Repeated sampling can scale coverage when candidates are verifiable \citep{brown2024monkeys}; s1 makes budget forcing an explicit intervention on thinking length \citep{s12025}; and parallel-distill-refine, TTRL, and MiGrATe make the deployment-time selector, verifier, or mixed policy part of the reported data object \citep{khatri2025scalerl,zuo2025ttrl,migrate2025}. Long-context reasoning adds another topology: more context can raise the ceiling while reducing efficiency, whereas Markovian Thinking changes the environment so reasoning proceeds through fixed-size chunks and learned state \citep{khatri2025scalerl,aghajohari2025markovian}. A pass@N score under serial extension, repeated sampling, parallel refinement, test-time training, or Markovian chunking is therefore not the same observable.

\subsection{Verifier Scaling and Release Lineage}
\label{subsec:verifier-scaling}

The verifier scales with the policy, but not necessarily in the right direction. Master-key attacks, GSM-Symbolic perturbations, and verifier-robustness studies show that reward signals can be broad yet brittle: LLM judges may reward trigger tokens, math accuracy may collapse under surface changes, rule verifiers can false-reject, and learned verifiers can become hackable \citep{zhao2025masterrm,gsmsymbolic2024,huang2025verifierrobustness}. Process-verifier corpora follow the same logic: PRM800K, Math-Shepherd, OmegaPRM, and Qwen2.5-Math-PRM differ in whether supervision comes from human labels, Monte-Carlo scores, first-error localization, or judge consensus \citep{lightman2023verify,mathshepherd2023,omegaprm2024,qwen25mathprm2025}. CoVerRL and DeepSeekMath-V2 suggest co-evolving generators and verifiers, but co-evolution increases the need for versioning rather than removing it \citep{pan2026coverrl,deepseek2025v32}.

The same versioning problem appears in benchmark reporting. Scores can shift with seed and budget \citep{hochlehnert2025sober}; search agents can retrieve benchmark-adjacent answers during evaluation \citep{han2025searchcontamination}; and teacher traits can pass through data even when semantic cues are stripped \citep{cloud2025subliminal}. Release time is therefore not a neutral timestamp but a proxy for a versioned data object: prompt pools, verifier versions, search budgets, contamination audits, teacher lineages, evaluation protocols, and refresh cadences may all have changed. A useful scaling report should disclose these fields before attributing a score difference to data scale, optimizer efficiency, or model size.

\section{Conclusion}
\label{sec:conclusion}

Post-training reasoning data have become the hidden substrate of reasoning-model progress, and the stream of releases is accelerating. 
The evidence reviewed here suggests that the reusable unit is not a prompt--response pair, but a verifier-bearing feedback interface whose value depends on the verifier, base model, lineage, optimizer, scaffold, and inference budget. 
The central unresolved question is therefore attribution: when a model improves, which part of that interface changed? 
This primer helps the community move from reporting gains to making them inspectable, comparable, and testable.


\section*{Limitations}
\label{sec:limitations}

This primer is limited by the public evidence available for post-training reasoning data. Closed pipelines, proprietary data mixtures, and undocumented release practices are necessarily absent, and many public reports omit lineage cards, verifier versions, compute and inference budgets, and contamination audits. Our synthesis is also question-driven rather than a formal meta-analysis: we include work that exposes reasoning-data objects, feedback interfaces, construction recipes, or scaling surfaces, which may underrepresent work that reports only aggregate benchmark gains or uses non-public artifacts. In addition, the evidence base is heterogeneous, spanning peer-reviewed papers, arXiv preprints, technical reports, benchmarks, and model cards; we use these sources to document public recipes and release practices, not to establish causal rankings. Finally, we do not independently re-run training recipes, audit contamination, or validate every verifier. The taxonomy should therefore be read as an attribution and reporting framework whose fields will need updating as reasoning post-training expands to new multimodal, multilingual, agentic, and co-evolving verifier-generator settings.

\section*{Ethics Statement}
\label{sec:ethics}

This work releases no new data or models. Still, the practices surveyed involve privacy-bearing traces, expert rubrics, safety boundaries, agentic actions, and high-cost RL pipelines, making license, provenance, annotator welfare, verifier hacking, contamination, and compute disclosure central ethical requirements for future releases.



\bibliography{references}

@techreport{openai_o1_2024,
  author = {OpenAI},
  title = {Learning to reason with {LLMs}},
  year = {2024},
  institution = {Technical report},
}

@article{bigmath2025,
  author = {Alon Albalak and Duy Phung and Nathan Lile and Rafael Rafailov and Kanishk Gandhi and Louis Castricato and Anikait Singh and Chase Blagden and Violet Xiang and Dakota Mahan and Nick Haber},
  title = {Big-Math: A Large-Scale, High-Quality Math Dataset for Reinforcement Learning in Language Models},
  journal = {arXiv preprint arXiv:2502.17387},
  year = {2025},
  eprint = {2502.17387},
  archivePrefix = {arXiv},
  primaryClass = {cs.LG},
  doi = {10.48550/arXiv.2502.17387},
  url = {https://arxiv.org/abs/2502.17387},
}

@article{healthbench2025,
  author = {Rahul K. Arora and Jason Wei and Rebecca Soskin Hicks and Preston Bowman and Joaquin Qui{\~n}onero-Candela and Foivos Tsimpourlas and Michael Sharman and Meghan Shah and Andrea Vallone and Alex Beutel and Johannes Heidecke and Karan Singhal},
  title = {HealthBench: Evaluating Large Language Models Towards Improved Human Health},
  journal = {arXiv preprint arXiv:2505.08775},
  year = {2025},
  eprint = {2505.08775},
  archivePrefix = {arXiv},
  primaryClass = {cs.CL},
  doi = {10.48550/arXiv.2505.08775},
  url = {https://arxiv.org/abs/2505.08775},
}

@article{hhrlhf2022,
  author = {Yuntao Bai and Andy Jones and Kamal Ndousse and Amanda Askell and Anna Chen and Nova DasSarma and Dawn Drain and Stanislav Fort and Deep Ganguli and Tom Henighan and Nicholas Joseph and Saurav Kadavath and Jackson Kernion and Tom Conerly and Sheer El-Showk and Nelson Elhage and Zac Hatfield-Dodds and Danny Hernandez and Tristan Hume and Scott Johnston and Shauna Kravec and Liane Lovitt and Neel Nanda and Catherine Olsson and Dario Amodei and Tom Brown and Jack Clark and Sam McCandlish and Chris Olah and Ben Mann and Jared Kaplan},
  title = {Training a Helpful and Harmless Assistant with Reinforcement Learning from Human Feedback},
  journal = {arXiv preprint arXiv:2204.05862},
  year = {2022},
  eprint = {2204.05862},
  archivePrefix = {arXiv},
  primaryClass = {cs.CL},
  doi = {10.48550/arXiv.2204.05862},
  url = {https://arxiv.org/abs/2204.05862},
}

@article{multiagentevolve2025,
  author = {Yixing Chen and Yiding Wang and Siqi Zhu and Haofei Yu and Tao Feng and Muhan Zhang and Mostofa Patwary and Jiaxuan You},
  title = {Multi-Agent Evolve: LLM Self-Improve through Co-evolution},
  journal = {arXiv preprint arXiv:2510.23595},
  year = {2025},
  eprint = {2510.23595},
  archivePrefix = {arXiv},
  primaryClass = {cs.AI},
  doi = {10.48550/arXiv.2510.23595},
  url = {https://arxiv.org/abs/2510.23595},
}

@article{frontiermath2024,
  author = {Elliot Glazer and Ege Erdil and Tamay Besiroglu and Diego Chicharro and Evan Chen and Alex Gunning and Caroline Falkman Olsson and Jean-Stanislas Denain and Anson Ho and Emily de Oliveira Santos and Olli J{\"a}rviniemi and Matthew Barnett and Robert Sandler and Matej Vrzala and Jaime Sevilla and Qiuyu Ren and Elizabeth Pratt and Lionel Levine and Grant Barkley and Natalie Stewart and Bogdan Grechuk and Tetiana Grechuk and Shreepranav Varma Enugandla and Mark Wildon},
  title = {FrontierMath: A Benchmark for Evaluating Advanced Mathematical Reasoning in AI},
  journal = {arXiv preprint arXiv:2411.04872},
  year = {2025},
  eprint = {2411.04872},
  archivePrefix = {arXiv},
  primaryClass = {cs.AI},
  doi = {10.48550/arXiv.2411.04872},
  url = {https://arxiv.org/abs/2411.04872},
}

@article{legalbench2023,
  author = {Neel Guha and Julian Nyarko and Daniel E. Ho and Christopher R{\'e} and Adam Chilton and Aditya Narayana and Alex Chohlas-Wood and Austin Peters and Brandon Waldon and Daniel N. Rockmore and Diego Zambrano and Dmitry Talisman and Enam Hoque and Faiz Surani and Frank Fagan and Galit Sarfaty and Gregory M. Dickinson and Haggai Porat and Jason Hegland and Jessica Wu and Joe Nudell and Joel Niklaus and John Nay and Jonathan H. Choi and Kevin Tobia and Margaret Hagan and Megan Ma and Michael Livermore and Nikon Rasumov-Rahe and Nils Holzenberger and Noam Kolt and Peter Henderson and Sean Rehaag and Sharad Goel and Shang Gao and Spencer Williams and Sunny Gandhi and Tom Zur and Varun Iyer and Zehua Li},
  title = {LegalBench: A Collaboratively Built Benchmark for Measuring Legal Reasoning in Large Language Models},
  journal = {arXiv preprint arXiv:2308.11462},
  year = {2023},
  eprint = {2308.11462},
  archivePrefix = {arXiv},
  primaryClass = {cs.CL},
  doi = {10.48550/arXiv.2308.11462},
  url = {https://arxiv.org/abs/2308.11462},
}

@article{rar2025,
  author = {Anisha Gunjal and Anthony Wang and Elaine Lau and Vaskar Nath and Yunzhong He and Bing Liu and Sean Hendryx},
  title = {Rubrics as Rewards: Reinforcement Learning Beyond Verifiable Domains},
  journal = {arXiv preprint arXiv:2507.17746},
  year = {2025},
  eprint = {2507.17746},
  archivePrefix = {arXiv},
  primaryClass = {cs.LG},
  doi = {10.48550/arXiv.2507.17746},
  url = {https://arxiv.org/abs/2507.17746},
}

@article{deepseekr12025,
  author = {DeepSeek-AI and Daya Guo and Dejian Yang and Haowei Zhang and Junxiao Song and Peiyi Wang and Qihao Zhu and Runxin Xu and Ruoyu Zhang and Shirong Ma and Xiao Bi and Xiaokang Zhang and Xingkai Yu and Yu Wu and Z. F. Wu and Zhibin Gou and Zhihong Shao and Zhuoshu Li and Ziyi Gao and Aixin Liu and Bing Xue and Bingxuan Wang and Bochao Wu and Bei Feng and Chengda Lu and Chenggang Zhao and Chengqi Deng and Chenyu Zhang and Chong Ruan and Damai Dai and Deli Chen and Dongjie Ji and Erhang Li and Fangyun Lin and Fucong Dai and Fuli Luo and Guangbo Hao and Guanting Chen and Guowei Li and H. Zhang and Han Bao and Hanwei Xu and Haocheng Wang and Honghui Ding and Huajian Xin and Huazuo Gao and Hui Qu and Hui Li and Jianzhong Guo and Jiashi Li and Jiawei Wang and Jingchang Chen and Jingyang Yuan and Junjie Qiu and Junlong Li and J. L. Cai and Jiaqi Ni and Jian Liang and Jin Chen and Kai Dong and Kai Hu and Kaige Gao and Kang Guan and Kexin Huang and Kuai Yu and Lean Wang and Lecong Zhang and Liang Zhao and Litong Wang and Liyue Zhang and Lei Xu and Leyi Xia and Mingchuan Zhang and Minghua Zhang and Minghui Tang and Meng Li and Miaojun Wang and Mingming Li and Ning Tian and Panpan Huang and Peng Zhang and Qiancheng Wang and Qinyu Chen and Qiushi Du and Ruiqi Ge and Ruisong Zhang and Ruizhe Pan and Runji Wang and R. J. Chen and R. L. Jin and Ruyi Chen and Shanghao Lu and Shangyan Zhou and Shanhuang Chen and Shengfeng Ye and Shiyu Wang and Shuiping Yu and Shunfeng Zhou and Shuting Pan and S. S. Li and Shuang Zhou and Shaoqing Wu and Shengfeng Ye and Tao Yun and Tian Pei and Tianyu Sun and T. Wang and Wangding Zeng and Wanjia Zhao and Wen Liu and Wenfeng Liang and Wenjun Gao and Wenqin Yu and Wentao Zhang and W. L. Xiao and Wei An and Xiaodong Liu and Xiaohan Wang and Xiaokang Chen and Xiaotao Nie and Xin Cheng and Xin Liu and Xin Xie and Xingchao Liu and Xinyu Yang and Xinyuan Li and Xuecheng Su and Xuheng Lin and X. Q. Li and Xiangyue Jin and Xiaojin Shen and Xiaosha Chen and Xiaowen Sun and Xiaoxiang Wang and Xinnan Song and Xinyi Zhou and Xianzu Wang and Xinxia Shan and Y. K. Li and Y. Q. Wang and Y. X. Wei and Yang Zhang and Yanhong Xu and Yao Li and Yao Zhao and Yaofeng Sun and Yaohui Wang and Yi Yu and Yichao Zhang and Yifan Shi and Yiliang Xiong and Ying He and Yishi Piao and Yisong Wang and Yixuan Tan and Yiyang Ma and Yiyuan Liu and Yongqiang Guo and Yuan Ou and Yuduan Wang and Yue Gong and Yuheng Zou and Yujia He and Yunfan Xiong and Yuxiang Luo and Yuxiang You and Yuxuan Liu and Yuyang Zhou and Y. X. Zhu and Yanhong Xu and Yanping Huang and Yaohui Li and Yi Zheng and Yuchen Zhu and Yunxian Ma and Ying Tang and Yukun Zha and Yuting Yan and Z. Z. Ren and Zehui Ren and Zhangli Sha and Zhe Fu and Zhean Xu and Zhenda Xie and Zhengyan Zhang and Zhewen Hao and Zhicheng Ma and Zhigang Yan and Zhiyu Wu and Zihui Gu and Zijia Zhu and Zijun Liu and Zilin Li and Ziwei Xie and Ziyang Song and Zizheng Pan and Zhen Huang and Zhipeng Xu and Zhongyu Zhang and Zhen Zhang},
  title = {DeepSeek-R1: Incentivizing Reasoning Capability in LLMs via Reinforcement Learning},
  journal = {arXiv preprint arXiv:2501.12948},
  year = {2025},
  eprint = {2501.12948},
  archivePrefix = {arXiv},
  primaryClass = {cs.CL},
  doi = {10.1038/s41586-025-09422-z},
  url = {https://arxiv.org/abs/2501.12948},
}

@article{wildguard2024,
  author = {Seungju Han and Kavel Rao and Allyson Ettinger and Liwei Jiang and Bill Yuchen Lin and Nathan Lambert and Yejin Choi and Nouha Dziri},
  title = {WildGuard: Open One-Stop Moderation Tools for Safety Risks, Jailbreaks, and Refusals of LLMs},
  journal = {arXiv preprint arXiv:2406.18495},
  year = {2024},
  eprint = {2406.18495},
  archivePrefix = {arXiv},
  primaryClass = {cs.CL},
  doi = {10.48550/arXiv.2406.18495},
  url = {https://arxiv.org/abs/2406.18495},
}

@article{r2egym2025,
  author = {Naman Jain and Jaskirat Singh and Manish Shetty and Liang Zheng and Koushik Sen and Ion Stoica},
  title = {R2E-Gym: Procedural Environments and Hybrid Verifiers for Scaling Open-Weights SWE Agents},
  journal = {arXiv preprint arXiv:2504.07164},
  year = {2025},
  eprint = {2504.07164},
  archivePrefix = {arXiv},
  primaryClass = {cs.SE},
  doi = {10.48550/arXiv.2504.07164},
  url = {https://arxiv.org/abs/2504.07164},
}

@article{kimi2025,
  author = {Kimi Team and Angang Du and Bofei Gao and Bowei Xing and Changjiu Jiang and Cheng Chen and Cheng Li and Chenjun Xiao and Chenzhuang Du and Chonghua Liao and Chuning Tang and Congcong Wang and Dehao Zhang and Enming Yuan and Enzhe Lu and Fengxiang Tang and Flood Sung and Guangda Wei and Guokun Lai and Haiqing Guo and Han Zhu and Hao Ding and Hao Hu and Hao Yang and Hao Zhang and Haotian Yao and Haotian Zhao and Haoyu Lu and Haoze Li and Haozhen Yu and Hongcheng Gao and Huabin Zheng and Huan Yuan and Jia Chen and Jianhang Guo and Jianlin Su and Jianzhou Wang and Jie Zhao and Jin Zhang and Jingyuan Liu and Junjie Yan and Junyan Wu and Lidong Shi and Ling Ye and Longhui Yu and Mengnan Dong and Neo Zhang and Ningchen Ma and Qiwei Pan and Qucheng Gong and Shaowei Liu and Shengling Ma and Shupeng Wei and Sihan Cao and Siying Huang and Tao Jiang and Weihao Gao and Weimin Xiong and Weiran He and Weixiao Huang and Weixin Xu and Wenhao Wu and Wenyang He and Xianghui Wei and Xianqing Jia and Xingzhe Wu and Xinran Xu and Xinxing Zu and Xinyu Zhou and Xuehai Pan and Y. Charles and Yang Li and Yangyang Hu and Yangyang Liu and Yanru Chen and Yejie Wang and Yibo Liu and Yidao Qin and Yifeng Liu and Ying Yang and Yiping Bao and Yulun Du and Yuxin Wu and Yuzhi Wang and Zaida Zhou and Zhaoji Wang and Zhaowei Li and Zhen Zhu and Zheng Zhang and Zhexu Wang and Zhilin Yang and Zhiqi Huang and Zihao Huang and Ziyao Xu and Zonghan Yang and Zongyu Lin},
  title = {Kimi k1.5: Scaling Reinforcement Learning with LLMs},
  journal = {arXiv preprint arXiv:2501.12599},
  year = {2025},
  eprint = {2501.12599},
  archivePrefix = {arXiv},
  primaryClass = {cs.AI},
  doi = {10.48550/arXiv.2501.12599},
  url = {https://arxiv.org/abs/2501.12599},
}

@article{lightman2023verify,
  author = {Hunter Lightman and Vineet Kosaraju and Yura Burda and Harri Edwards and Bowen Baker and Teddy Lee and Jan Leike and John Schulman and Ilya Sutskever and Karl Cobbe},
  title = {Let's Verify Step by Step},
  journal = {arXiv preprint arXiv:2305.20050},
  year = {2023},
  eprint = {2305.20050},
  archivePrefix = {arXiv},
  primaryClass = {cs.LG},
  doi = {10.48550/arXiv.2305.20050},
  url = {https://arxiv.org/abs/2305.20050},
}

@article{truthfulqa2022,
  author = {Stephanie Lin and Jacob Hilton and Owain Evans},
  title = {TruthfulQA: Measuring How Models Mimic Human Falsehoods},
  journal = {arXiv preprint arXiv:2109.07958},
  year = {2022},
  eprint = {2109.07958},
  archivePrefix = {arXiv},
  primaryClass = {cs.CL},
  doi = {10.48550/arXiv.2109.07958},
  url = {https://arxiv.org/abs/2109.07958},
}

@article{omegaprm2024,
  author = {Liangchen Luo and Yinxiao Liu and Rosanne Liu and Samrat Phatale and Meiqi Guo and Harsh Lara and Yunxuan Li and Lei Shu and Yun Zhu and Lei Meng and Jiao Sun and Abhinav Rastogi},
  title = {Improve Mathematical Reasoning in Language Models by Automated Process Supervision},
  journal = {arXiv preprint arXiv:2406.06592},
  year = {2024},
  eprint = {2406.06592},
  archivePrefix = {arXiv},
  primaryClass = {cs.CL},
  doi = {10.48550/arXiv.2406.06592},
  url = {https://arxiv.org/abs/2406.06592},
}

@article{harmbench2024,
  author = {Mantas Mazeika and Long Phan and Xuwang Yin and Andy Zou and Zifan Wang and Norman Mu and Elham Sakhaee and Nathaniel Li and Steven Basart and Bo Li and David Forsyth and Dan Hendrycks},
  title = {HarmBench: A Standardized Evaluation Framework for Automated Red Teaming and Robust Refusal},
  journal = {arXiv preprint arXiv:2402.04249},
  year = {2024},
  eprint = {2402.04249},
  archivePrefix = {arXiv},
  primaryClass = {cs.LG},
  doi = {10.48550/arXiv.2402.04249},
  url = {https://arxiv.org/abs/2402.04249},
}

@article{gsmsymbolic2024,
  author = {Iman Mirzadeh and Keivan Alizadeh and Hooman Shahrokhi and Oncel Tuzel and Samy Bengio and Mehrdad Farajtabar},
  title = {GSM-Symbolic: Understanding the Limitations of Mathematical Reasoning in Large Language Models},
  journal = {arXiv preprint arXiv:2410.05229},
  year = {2025},
  eprint = {2410.05229},
  archivePrefix = {arXiv},
  primaryClass = {cs.LG},
  doi = {10.48550/arXiv.2410.05229},
  url = {https://arxiv.org/abs/2410.05229},
}

@article{openmathreasoning2025,
  author = {Ivan Moshkov and Darragh Hanley and Ivan Sorokin and Shubham Toshniwal and Christof Henkel and Benedikt Schifferer and Wei Du and Igor Gitman},
  title = {AIMO-2 Winning Solution: Building State-of-the-Art Mathematical Reasoning Models with OpenMathReasoning dataset},
  journal = {arXiv preprint arXiv:2504.16891},
  year = {2025},
  eprint = {2504.16891},
  archivePrefix = {arXiv},
  primaryClass = {cs.AI},
  doi = {10.48550/arXiv.2504.16891},
  url = {https://arxiv.org/abs/2504.16891},
}

@article{s12025,
  author = {Niklas Muennighoff and Zitong Yang and Weijia Shi and Xiang Lisa Li and Li Fei-Fei and Hannaneh Hajishirzi and Luke Zettlemoyer and Percy Liang and Emmanuel Cand{\`e}s and Tatsunori Hashimoto},
  title = {s1: Simple test-time scaling},
  journal = {arXiv preprint arXiv:2501.19393},
  year = {2025},
  eprint = {2501.19393},
  archivePrefix = {arXiv},
  primaryClass = {cs.CL},
  doi = {10.48550/arXiv.2501.19393},
  url = {https://arxiv.org/abs/2501.19393},
}

@article{swegym2024,
  author = {Jiayi Pan and Xingyao Wang and Graham Neubig and Navdeep Jaitly and Heng Ji and Alane Suhr and Yizhe Zhang},
  title = {Training Software Engineering Agents and Verifiers with SWE-Gym},
  journal = {arXiv preprint arXiv:2412.21139},
  year = {2025},
  eprint = {2412.21139},
  archivePrefix = {arXiv},
  primaryClass = {cs.SE},
  doi = {10.48550/arXiv.2412.21139},
  url = {https://arxiv.org/abs/2412.21139},
}

@article{gorilla2023,
  author = {Shishir G. Patil and Tianjun Zhang and Xin Wang and Joseph E. Gonzalez},
  title = {Gorilla: Large Language Model Connected with Massive APIs},
  journal = {arXiv preprint arXiv:2305.15334},
  year = {2023},
  eprint = {2305.15334},
  archivePrefix = {arXiv},
  primaryClass = {cs.CL},
  doi = {10.48550/arXiv.2305.15334},
  url = {https://arxiv.org/abs/2305.15334},
}

@article{hle2025,
  author = {Long Phan and Alice Gatti and Ziwen Han and Nathaniel Li and Josephina Hu and Hugh Zhang and Chen Bo Calvin Zhang and Mohamed Shaaban and John Ling and Sean Shi and Michael Choi and Anish Agrawal and Arnav Chopra and Adam Khoja and Ryan Kim and Richard Ren and Jason Hausenloy and Oliver Zhang and Mantas Mazeika and Dmitry Dodonov and Tung Nguyen and Jaeho Lee and Daron Anderson and Mikhail Doroshenko and Alun Cennyth Stokes and Mobeen Mahmood and Oleksandr Pokutnyi and Oleg Iskra and Jessica P. Wang and John-Clark Levin and Mstyslav Kazakov and Fiona Feng and Steven Y. Feng and Haoran Zhao and Michael Yu and Varun Gangal and Chelsea Zou and Zihan Wang and Serguei Popov and Robert Gerbicz and Geoff Galgon and Johannes Schmitt and Will Yeadon and Yongki Lee and Scott Sauers and Alvaro Sanchez and Fabian Giska and Marc Roth and S{\o}ren Riis and Saiteja Utpala and Noah Burns and Gashaw M. Goshu and Mohinder Maheshbhai Naiya and Chidozie Agu and Zachary Giboney and Antrell Cheatom and Francesco Fournier-Facio and Sarah-Jane Crowson and Lennart Finke and Zerui Cheng and Jennifer Zampese and Ryan G. Hoerr and Mark Nandor and Hyunwoo Park and Tim Gehrunger and Jiaqi Cai and Ben McCarty and Alexis C Garretson and Edwin Taylor and Damien Sileo and Qiuyu Ren and Usman Qazi and Lianghui Li and Jungbae Nam and John B. Wydallis and Pavel Arkhipov and Jack Wei Lun Shi and Aras Bacho and Chris G. Willcocks and Hangrui Cao and Sumeet Motwani and Emily de Oliveira Santos and Johannes Veith and Edward Vendrow and Doru Cojoc and Kengo Zenitani and Joshua Robinson and Longke Tang and Yuqi Li and Joshua Vendrow and Natanael Wildner Fraga and Vladyslav Kuchkin and Andrey Pupasov Maksimov and Pierre Marion and Denis Efremov and Jayson Lynch and Kaiqu Liang and Aleksandar Mikov and Andrew Gritsevskiy and Julien Guillod and G{\"o}zdenur Demir and Dakotah Martinez and Ben Pageler and Kevin Zhou and Saeed Soori and Ori Press and Henry Tang and Paolo Rissone and Sean R. Green and Lina Br{\"u}ssel and Moon Twayana and Aymeric Dieuleveut and Joseph Marvin Imperial and Ameya Prabhu and Jinzhou Yang and Nick Crispino and Arun Rao and Dimitri Zvonkine and Gabriel Loiseau and Mikhail Kalinin and Marco Lukas and Ciprian Manolescu and Nate Stambaugh and Subrata Mishra and Tad Hogg and Carlo Bosio and Brian P Coppola and Julian Salazar and Jaehyeok Jin and Rafael Sayous and Stefan Ivanov and Philippe Schwaller and Shaipranesh Senthilkuma and Andres M Bran and Andres Algaba and Kelsey Van den Houte and Lynn Van Der Sypt and Brecht Verbeken and David Noever and Alexei Kopylov and Benjamin Myklebust and Bikun Li and Lisa Schut and Evgenii Zheltonozhskii and Qiaochu Yuan and Derek Lim and Richard Stanley and Tong Yang and John Maar and Julian Wykowski and Mart{\'i} Oller and Anmol Sahu and Cesare Giulio Ardito and Yuzheng Hu and Ariel Ghislain Kemogne Kamdoum and Alvin Jin and Tobias Garcia Vilchis and Yuexuan Zu and Martin Lackner and James Koppel and Gongbo Sun and Daniil S. Antonenko and Steffi Chern and Bingchen Zhao and Pierrot Arsene and Joseph M Cavanagh and Daofeng Li and Jiawei Shen and Donato Crisostomi and Wenjin Zhang and Ali Dehghan and Sergey Ivanov and David Perrella and Nurdin Kaparov and Allen Zang and Ilia Sucholutsky and Arina Kharlamova and Daniil Orel and Vladislav Poritski and Shalev Ben-David and Zachary Berger and Parker Whitfill and Michael Foster and Daniel Munro and Linh Ho and Shankar Sivarajan and Dan Bar Hava and Aleksey Kuchkin and David Holmes and Alexandra Rodriguez-Romero and Frank Sommerhage and Anji Zhang and Richard Moat and Keith Schneider and Zakayo Kazibwe and Don Clarke and Dae Hyun Kim and Felipe Meneguitti Dias and Sara Fish and Veit Elser and Tobias Kreiman and Victor Efren Guadarrama Vilchis and Immo Klose and Ujjwala Anantheswaran and Adam Zweiger and Kaivalya Rawal and Jeffery Li and Jeremy Nguyen and Nicolas Daans and Haline Heidinger and Maksim Radionov and V{\'a}clav Rozho{\v{n}} and Vincent Ginis and Christian Stump and Niv Cohen and Rafa{\l} Po{\'s}wiata and Josef Tkadlec and Alan Goldfarb and Chenguang Wang and Piotr Padlewski and Stanislaw Barzowski and Kyle Montgomery and Ryan Stendall and Jamie Tucker-Foltz and Jack Stade and T. Ryan Rogers and Tom Goertzen and Declan Grabb and Abhishek Shukla and Alan Givr{\'e} and John Arnold Ambay and Archan Sen and Muhammad Fayez Aziz and Mark H Inlow and Hao He and Ling Zhang and Younesse Kaddar and Ivar {\"A}ngquist and Yanxu Chen and Harrison K Wang and Kalyan Ramakrishnan and Elliott Thornley and Antonio Terpin and Hailey Schoelkopf and Eric Zheng and Avishy Carmi and Ethan D. L. Brown and Kelin Zhu and Max Bartolo and Richard Wheeler and Martin Stehberger and Peter Bradshaw and JP Heimonen and Kaustubh Sridhar and Ido Akov and Jennifer Sandlin and Yury Makarychev and Joanna Tam and Hieu Hoang and David M. Cunningham and Vladimir Goryachev and Demosthenes Patramanis and Michael Krause and Andrew Redenti and David Aldous and Jesyin Lai and Shannon Coleman and Jiangnan Xu and Sangwon Lee and Ilias Magoulas and Sandy Zhao and Ning Tang and Michael K. Cohen and Orr Paradise and Jan Hendrik Kirchner and Maksym Ovchynnikov and Jason O. Matos and Adithya Shenoy and Michael Wang and Yuzhou Nie and Anna Sztyber-Betley and Paolo Faraboschi and Robin Riblet and Jonathan Crozier and Shiv Halasyamani and Shreyas Verma and Prashant Joshi and Eli Meril and Ziqiao Ma and J{\'e}r{\'e}my Andr{\'e}oletti and Raghav Singhal and Jacob Platnick and Volodymyr Nevirkovets and Luke Basler and Alexander Ivanov and Seri Khoury and Nils Gustafsson and Marco Piccardo and Hamid Mostaghimi and Qijia Chen and Virendra Singh and Tran Quoc Kh{\'a}nh and Paul Rosu and Hannah Szlyk and Zachary Brown and Himanshu Narayan and Aline Menezes and Jonathan Roberts and William Alley and Kunyang Sun and Arkil Patel and Max Lamparth and Anka Reuel and Linwei Xin and Hanmeng Xu and Jacob Loader and Freddie Martin and Zixuan Wang and Andrea Achilleos and Thomas Preu and Tomek Korbak and Ida Bosio and Fereshteh Kazemi and Ziye Chen and Bir{\'o} B{\'a}lint and Eve J. Y. Lo and Jiaqi Wang and Maria In{\^e}s S. Nunes and Jeremiah Milbauer and M Saiful Bari and Zihao Wang and Behzad Ansarinejad and Yewen Sun and Stephane Durand and Hossam Elgnainy and Guillaume Douville and Daniel Tordera and George Balabanian and Hew Wolff and Lynna Kvistad and Hsiaoyun Milliron and Ahmad Sakor and Murat Eron and Andrew Favre D. O. and Shailesh Shah and Xiaoxiang Zhou and Firuz Kamalov and Sherwin Abdoli and Tim Santens and Shaul Barkan and Allison Tee and Robin Zhang and Alessandro Tomasiello and G. Bruno De Luca and Shi-Zhuo Looi and Vinh-Kha Le and Noam Kolt and Jiayi Pan and Emma Rodman and Jacob Drori and Carl J Fossum and Niklas Muennighoff and Milind Jagota and Ronak Pradeep and Honglu Fan and Jonathan Eicher and Michael Chen and Kushal Thaman and William Merrill and Moritz Firsching and Carter Harris and Stefan Ciob{\^a}c{\u{a}} and Jason Gross and Rohan Pandey and Ilya Gusev and Adam Jones and Shashank Agnihotri and Pavel Zhelnov and Mohammadreza Mofayezi and Alexander Piperski and David K. Zhang and Kostiantyn Dobarskyi and Roman Leventov and Ignat Soroko and Joshua Duersch and Vage Taamazyan and Andrew Ho and Wenjie Ma and William Held and Ruicheng Xian and Armel Randy Zebaze and Mohanad Mohamed and Julian Noah Leser and Michelle X Yuan and Laila Yacar and Johannes Lengler and Katarzyna Olszewska and Claudio Di Fratta and Edson Oliveira and Joseph W. Jackson and Andy Zou and Muthu Chidambaram and Timothy Manik and Hector Haffenden and Dashiell Stander and Ali Dasouqi and Alexander Shen and Bita Golshani and David Stap and Egor Kretov and Mikalai Uzhou and Alina Borisovna Zhidkovskaya and Nick Winter and Miguel Orbegozo Rodriguez and Robert Lauff and Dustin Wehr and Colin Tang and Zaki Hossain and Shaun Phillips and Fortuna Samuele and Fredrik Ekstr{\"o}m and Angela Hammon and Oam Patel and Faraz Farhidi and George Medley and Forough Mohammadzadeh and Madellene Pe{\~n}aflor and Haile Kassahun and Alena Friedrich and Rayner Hernandez Perez and Daniel Pyda and Taom Sakal and Omkar Dhamane and Ali Khajegili Mirabadi and Eric Hallman and Kenchi Okutsu and Mike Battaglia and Mohammad Maghsoudimehrabani and Alon Amit and Dave Hulbert and Roberto Pereira and Simon Weber and Handoko and Anton Peristyy and Stephen Malina and Mustafa Mehkary and Rami Aly and Frank Reidegeld and Anna-Katharina Dick and Cary Friday and Mukhwinder Singh and Hassan Shapourian and Wanyoung Kim and Mariana Costa and Hubeyb Gurdogan and Harsh Kumar and Chiara Ceconello and Chao Zhuang and Haon Park and Micah Carroll and Andrew R. Tawfeek and Stefan Steinerberger and Daattavya Aggarwal and Michael Kirchhof and Linjie Dai and Evan Kim and Johan Ferret and Jainam Shah and Yuzhou Wang and Minghao Yan and Krzysztof Burdzy and Lixin Zhang and Antonio Franca and Diana T. Pham and Kang Yong Loh and Joshua Robinson and Abram Jackson and Paolo Giordano and Philipp Petersen and Adrian Cosma and Jesus Colino and Colin White and Jacob Votava and Vladimir Vinnikov and Ethan Delaney and Petr Spelda and Vit Stritecky and Syed M. Shahid and Jean-Christophe Mourrat and Lavr Vetoshkin and Koen Sponselee and Renas Bacho and Zheng-Xin Yong and Florencia de la Rosa and Nathan Cho and Xiuyu Li and Guillaume Malod and Orion Weller and Guglielmo Albani and Leon Lang and Julien Laurendeau and Dmitry Kazakov and Fatimah Adesanya and Julien Portier and Lawrence Hollom and Victor Souza and Yuchen Anna Zhou and Julien Degorre and Yi{\u{g}}it Yal{\i}n and Gbenga Daniel Obikoya and Rai and Filippo Bigi and M. C. Bosc{\'a} and Oleg Shumar and Kaniuar Bacho and Gabriel Recchia and Mara Popescu and Nikita Shulga and Ngefor Mildred Tanwie and Thomas C. H. Lux and Ben Rank and Colin Ni and Matthew Brooks and Alesia Yakimchyk and Huanxu and Liu and Stefano Cavalleri and Olle H{\"a}ggstr{\"o}m and Emil Verkama and Joshua Newbould and Hans Gundlach and Leonor Brito-Santana and Brian Amaro and Vivek Vajipey and Rynaa Grover and Ting Wang and Yosi Kratish and Wen-Ding Li and Sivakanth Gopi and Andrea Caciolai and Christian Schroeder de Witt and Pablo Hern{\'a}ndez-C{\'a}mara and Emanuele Rodol{\`a} and Jules Robins and Dominic Williamson and Vincent Cheng and Brad Raynor and Hao Qi and Ben Segev and Jingxuan Fan and Sarah Martinson and Erik Y. Wang and Kaylie Hausknecht and Michael P. Brenner and Mao Mao and Christoph Demian and Peyman Kassani and Xinyu Zhang and David Avagian and Eshawn Jessica Scipio and Alon Ragoler and Justin Tan and Blake Sims and Rebeka Plecnik and Aaron Kirtland and Omer Faruk Bodur and D. P. Shinde and Yan Carlos Leyva Labrador and Zahra Adoul and Mohamed Zekry and Ali Karakoc and Tania C. B. Santos and Samir Shamseldeen and Loukmane Karim and Anna Liakhovitskaia and Nate Resman and Nicholas Farina and Juan Carlos Gonzalez and Gabe Maayan and Earth Anderson and Rodrigo De Oliveira Pena and Elizabeth Kelley and Hodjat Mariji and Rasoul Pouriamanesh and Wentao Wu and Ross Finocchio and Ismail Alarab and Joshua Cole and Danyelle Ferreira and Bryan Johnson and Mohammad Safdari and Liangti Dai and Siriphan Arthornthurasuk and Isaac C. McAlister and Alejandro Jos{\'e} Moyano and Alexey Pronin and Jing Fan and Angel Ramirez-Trinidad and Yana Malysheva and Daphiny Pottmaier and Omid Taheri and Stanley Stepanic and Samuel Perry and Luke Askew and Ra{\'u}l Adri{\'a}n Huerta Rodr{\'i}guez and Ali M. R. Minissi and Ricardo Lorena and Krishnamurthy Iyer and Arshad Anil Fasiludeen and Ronald Clark and Josh Ducey and Matheus Piza and Maja Somrak and Eric Vergo and Juehang Qin and Benj{\'a}min Borb{\'a}s and Eric Chu and Jack Lindsey and Antoine Jallon and I. M. J. McInnis and Evan Chen and Avi Semler and Luk Gloor and Tej Shah and Marc Carauleanu and Pascal Lauer and Tran {\DJ}uc Huy and Hossein Shahrtash and Emilien Duc and Lukas Lewark and Assaf Brown and Samuel Albanie and Brian Weber and Warren S. Vaz and Pierre Clavier and Yiyang Fan and Gabriel Poesia Reis e Silva and Long and Lian and Marcus Abramovitch and Xi Jiang and Sandra Mendoza and Murat Islam and Juan Gonzalez and Vasilios Mavroudis and Justin Xu and Pawan Kumar and Laxman Prasad Goswami and Daniel Bugas and Nasser Heydari and Ferenc Jeanplong and Thorben Jansen and Antonella Pinto and Archimedes Apronti and Abdallah Galal and Ng Ze-An and Ankit Singh and Tong Jiang and Joan of Arc Xavier and Kanu Priya Agarwal and Mohammed Berkani and Gang Zhang and Zhehang Du and Benedito Alves de Oliveira Junior and Dmitry Malishev and Nicolas Remy and Taylor D. Hartman and Tim Tarver and Stephen Mensah and Gautier Abou Loume and Wiktor Morak and Farzad Habibi and Sarah Hoback and Will Cai and Javier Gimenez and Roselynn Grace Montecillo and Jakub {\L}ucki and Russell Campbell and Asankhaya Sharma and Khalida Meer and Shreen Gul and Daniel Espinosa Gonzalez and Xavier Alapont and Alex Hoover and Gunjan Chhablani and Freddie Vargus and Arunim Agarwal and Yibo Jiang and Deepakkumar Patil and David Outevsky and Kevin Joseph Scaria and Rajat Maheshwari and Abdelkader Dendane and Priti Shukla and Ashley Cartwright and Sergei Bogdanov and Niels M{\"u}ndler and S{\"o}ren M{\"o}ller and Luca Arnaboldi and Kunvar Thaman and Muhammad Rehan Siddiqi and Prajvi Saxena and Himanshu Gupta and Tony Fruhauff and Glen Sherman and M{\'a}ty{\'a}s Vincze and Siranut Usawasutsakorn and Dylan Ler and Anil Radhakrishnan and Innocent Enyekwe and Sk Md Salauddin and Jiang Muzhen and Aleksandr Maksapetyan and Vivien Rossbach and Chris Harjadi and Mohsen Bahaloohoreh and Claire Sparrow and Jasdeep Sidhu and Sam Ali and Song Bian and John Lai and Eric Singer and Justine Leon Uro and Greg Bateman and Mohamed Sayed and Ahmed Menshawy and Darling Duclosel and Dario Bezzi and Yashaswini Jain and Ashley Aaron and Murat Tiryakioglu and Sheeshram Siddh and Keith Krenek and Imad Ali Shah and Jun Jin and Scott Creighton and Denis Peskoff and Zienab EL-Wasif and Ragavendran P V and Michael Richmond and Joseph McGowan and Tejal Patwardhan and Hao-Yu Sun and Ting Sun and Nikola Zubi{\'c} and Samuele Sala and Stephen Ebert and Jean Kaddour and Manuel Schottdorf and Dianzhuo Wang and Gerol Petruzella and Alex Meiburg and Tilen Medved and Ali ElSheikh and S Ashwin Hebbar and Lorenzo Vaquero and Xianjun Yang and Jason Poulos and Vil{\'e}m Zouhar and Sergey Bogdanik and Mingfang Zhang and Jorge Sanz-Ros and David Anugraha and Yinwei Dai and Anh N. Nhu and Xue Wang and Ali Anil Demircali and Zhibai Jia and Yuyin Zhou and Juncheng Wu and Mike He and Nitin Chandok and Aarush Sinha and Gaoxiang Luo and Long Le and Micka{\"e}l Noy{\'e} and Micha{\l} Pere{\l}kiewicz and Ioannis Pantidis and Tianbo Qi and Soham Sachin Purohit and Letitia Parcalabescu and Thai-Hoa Nguyen and Genta Indra Winata and Edoardo M. Ponti and Hanchen Li and Kaustubh Dhole and Jongee Park and Dario Abbondanza and Yuanli Wang and Anupam Nayak and Diogo M. Caetano and Antonio A. W. L. Wong and Maria del Rio-Chanona and D{\'a}niel Kondor and Pieter Francois and Ed Chalstrey and Jakob Zsambok and Dan Hoyer and Jenny Reddish and Jakob Hauser and Francisco-Javier Rodrigo-Gin{\'e}s and Suchandra Datta and Maxwell Shepherd and Thom Kamphuis and Qizheng Zhang and Hyunjun Kim and Ruiji Sun and Jianzhu Yao and Franck Dernoncourt and Satyapriya Krishna and Sina Rismanchian and Bonan Pu and Francesco Pinto and Yingheng Wang and Kumar Shridhar and Kalon J. Overholt and Glib Briia and Hieu Nguyen and David and Soler Bartomeu and Tony CY Pang and Adam Wecker and Yifan Xiong and Fanfei Li and Lukas S. Huber and Joshua Jaeger and Romano De Maddalena and Xing Han L{\`u} and Yuhui Zhang and Claas Beger and Patrick Tser Jern Kon and Sean Li and Vivek Sanker and Ming Yin and Yihao Liang and Xinlu Zhang and Ankit Agrawal and Li S. Yifei and Zechen Zhang and Mu Cai and Yasin Sonmez and Costin Cozianu and Changhao Li and Alex Slen and Shoubin Yu and Hyun Kyu Park and Gabriele Sarti and Marcin Bria{\'n}ski and Alessandro Stolfo and Truong An Nguyen and Mike Zhang and Yotam Perlitz and Jose Hernandez-Orallo and Runjia Li and Amin Shabani and Felix Juefei-Xu and Shikhar Dhingra and Orr Zohar and My Chiffon Nguyen and Alexander Pondaven and Abdurrahim Yilmaz and Xuandong Zhao and Chuanyang Jin and Muyan Jiang and Stefan Todoran and Xinyao Han and Jules Kreuer and Brian Rabern and Anna Plassart and Martino Maggetti and Luther Yap and Robert Geirhos and Jonathon Kean and Dingsu Wang and Sina Mollaei and Chenkai Sun and Yifan Yin and Shiqi Wang and Rui Li and Yaowen Chang and Anjiang Wei and Alice Bizeul and Xiaohan Wang and Alexandre Oliveira Arrais and Kushin Mukherjee and Jorge Chamorro-Padial and Jiachen Liu and Xingyu Qu and Junyi Guan and Adam Bouyamourn and Shuyu Wu and Martyna Plomecka and Junda Chen and Mengze Tang and Jiaqi Deng and Shreyas Subramanian and Haocheng Xi and Haoxuan Chen and Weizhi Zhang and Yinuo Ren and Haoqin Tu and Sejong Kim and Yushun Chen and Sara Vera Marjanovi{\'c} and Junwoo Ha and Grzegorz Luczyna and Jeff J. Ma and Zewen Shen and Dawn Song and Cedegao E. Zhang and Zhun Wang and Ga{\"e}l Gendron and Yunze Xiao and Leo Smucker and Erica Weng and Kwok Hao Lee and Zhe Ye and Stefano Ermon and Ignacio D. Lopez-Miguel and Theo Knights and Anthony Gitter and Namkyu Park and Boyi Wei and Hongzheng Chen and Kunal Pai and Ahmed Elkhanany and Han Lin and Philipp D. Siedler and Jichao Fang and Ritwik Mishra and K{\'a}roly Zsolnai-Feh{\'e}r and Xilin Jiang and Shadab Khan and Jun Yuan and Rishab Kumar Jain and Xi Lin and Mike Peterson and Zhe Wang and Aditya Malusare and Maosen Tang and Isha Gupta and Ivan Fosin and Timothy Kang and Barbara Dworakowska and Kazuki Matsumoto and Guangyao Zheng and Gerben Sewuster and Jorge Pretel Villanueva and Ivan Rannev and Igor Chernyavsky and Jiale Chen and Deepayan Banik and Ben Racz and Wenchao Dong and Jianxin Wang and Laila Bashmal and Duarte V. Gon{\c{c}}alves and Wei Hu and Kaushik Bar and Ondrej Bohdal and Atharv Singh Patlan and Shehzaad Dhuliawala and Caroline Geirhos and Julien Wist and Yuval Kansal and Bingsen Chen and Kutay Tire and Atak Talay Y{\"u}cel and Brandon Christof and Veerupaksh Singla and Zijian Song and Sanxing Chen and Jiaxin Ge and Kaustubh Ponkshe and Isaac Park and Tianneng Shi and Martin Q. Ma and Joshua Mak and Sherwin Lai and Antoine Moulin and Zhuo Cheng and Zhanda Zhu and Ziyi Zhang and Vaidehi Patil and Ketan Jha and Qiutong Men and Jiaxuan Wu and Tianchi Zhang and Bruno Hebling Vieira and Alham Fikri Aji and Jae-Won Chung and Mohammed Mahfoud and Ha Thi Hoang and Marc Sperzel and Wei Hao and Kristof Meding and Sihan Xu and Vassilis Kostakos and Davide Manini and Yueying Liu and Christopher Toukmaji and Jay Paek and Eunmi Yu and Arif Engin Demircali and Zhiyi Sun and Ivan Dewerpe and Hongsen Qin and Roman Pflugfelder and James Bailey and Johnathan Morris and Ville Heilala and Sybille Rosset and Zishun Yu and Peter E. Chen and Woongyeong Yeo and Eeshaan Jain and Ryan Yang and Sreekar Chigurupati and Julia Chernyavsky and Sai Prajwal Reddy and Subhashini Venugopalan and Hunar Batra and Core Francisco Park and Hieu Tran and Guilherme Maximiano and Genghan Zhang and Yizhuo Liang and Hu Shiyu and Rongwu Xu and Rui Pan and Siddharth Suresh and Ziqi Liu and Samaksh Gulati and Songyang Zhang and Peter Turchin and Christopher W. Bartlett and Christopher R. Scotese and Phuong M. Cao and Ben Wu and Jacek Karwowski and Davide Scaramuzza and Aakaash Nattanmai and Gordon McKellips and Anish Cheraku and Asim Suhail and Ethan Luo and Marvin Deng and Jason Luo and Ashley Zhang and Kavin Jindel and Jay Paek and Kasper Halevy and Allen Baranov and Michael Liu and Advaith Avadhanam and David Zhang and Vincent Cheng and Brad Ma and Evan Fu and Liam Do and Joshua Lass and Hubert Yang and Surya Sunkari and Vishruth Bharath and Violet Ai and James Leung and Rishit Agrawal and Alan Zhou and Kevin Chen and Tejas Kalpathi and Ziqi Xu and Gavin Wang and Tyler Xiao and Erik Maung and Sam Lee and Ryan Yang and Roy Yue and Ben Zhao and Julia Yoon and Sunny Sun and Aryan Singh and Ethan Luo and Clark Peng and Tyler Osbey and Taozhi Wang and Daryl Echeazu and Hubert Yang and Timothy Wu and Spandan Patel and Vidhi Kulkarni and Vijaykaarti Sundarapandiyan and Ashley Zhang and Andrew Le and Zafir Nasim and Srikar Yalam and Ritesh Kasamsetty and Soham Samal and Hubert Yang and David Sun and Nihar Shah and Abhijeet Saha and Alex Zhang and Leon Nguyen and Laasya Nagumalli and Kaixin Wang and Alan Zhou and Aidan Wu and Jason Luo and Anwith Telluri and Steven Dillmann and Zhengxiang Wang and Junyu Luo and Hugo Lunn and Artem Gazizov and Haoran Qiu and Allen G Hart and Rickard Br{\"u}el Gabrielsson and Ido Akov and Artem Lukoianov and Summer Yue and Alexandr Wang and Dan Hendrycks},
  title = {Humanity's Last Exam},
  journal = {arXiv preprint arXiv:2501.14249},
  year = {2026},
  eprint = {2501.14249},
  archivePrefix = {arXiv},
  primaryClass = {cs.LG},
  doi = {10.1038/s41586-025-09962-4},
  url = {https://arxiv.org/abs/2501.14249},
}

@article{toolllm2023,
  author = {Yujia Qin and Shihao Liang and Yining Ye and Kunlun Zhu and Lan Yan and Yaxi Lu and Yankai Lin and Xin Cong and Xiangru Tang and Bill Qian and Sihan Zhao and Lauren Hong and Runchu Tian and Ruobing Xie and Jie Zhou and Mark Gerstein and Dahai Li and Zhiyuan Liu and Maosong Sun},
  title = {ToolLLM: Facilitating Large Language Models to Master 16000+ Real-world APIs},
  journal = {arXiv preprint arXiv:2307.16789},
  year = {2023},
  eprint = {2307.16789},
  archivePrefix = {arXiv},
  primaryClass = {cs.AI},
  doi = {10.48550/arXiv.2307.16789},
  url = {https://arxiv.org/abs/2307.16789},
}

@article{gpqa2023,
  author = {David Rein and Betty Li Hou and Asa Cooper Stickland and Jackson Petty and Richard Yuanzhe Pang and Julien Dirani and Julian Michael and Samuel R. Bowman},
  title = {GPQA: A Graduate-Level Google-Proof Q\&A Benchmark},
  journal = {arXiv preprint arXiv:2311.12022},
  year = {2023},
  eprint = {2311.12022},
  archivePrefix = {arXiv},
  primaryClass = {cs.AI},
  doi = {10.48550/arXiv.2311.12022},
  url = {https://arxiv.org/abs/2311.12022},
}

@article{deepseekproverv2_2025,
  author = {Z. Z. Ren and Zhihong Shao and Junxiao Song and Huajian Xin and Haocheng Wang and Wanjia Zhao and Liyue Zhang and Zhe Fu and Qihao Zhu and Dejian Yang and Z. F. Wu and Zhibin Gou and Shirong Ma and Hongxuan Tang and Yuxuan Liu and Wenjun Gao and Daya Guo and Chong Ruan},
  title = {DeepSeek-Prover-V2: Advancing Formal Mathematical Reasoning via Reinforcement Learning for Subgoal Decomposition},
  journal = {arXiv preprint arXiv:2504.21801},
  year = {2025},
  eprint = {2504.21801},
  archivePrefix = {arXiv},
  primaryClass = {cs.CL},
  doi = {10.48550/arXiv.2504.21801},
  url = {https://arxiv.org/abs/2504.21801},
}

@article{xstest2024,
  author = {Paul R{\"o}ttger and Hannah Rose Kirk and Bertie Vidgen and Giuseppe Attanasio and Federico Bianchi and Dirk Hovy},
  title = {XSTest: A Test Suite for Identifying Exaggerated Safety Behaviours in Large Language Models},
  journal = {arXiv preprint arXiv:2308.01263},
  year = {2024},
  eprint = {2308.01263},
  archivePrefix = {arXiv},
  primaryClass = {cs.CL},
  doi = {10.48550/arXiv.2308.01263},
  url = {https://arxiv.org/abs/2308.01263},
}

@article{spuriousrewards2025,
  author = {Rulin Shao and Shuyue Stella Li and Rui Xin and Scott Geng and Yiping Wang and Sewoong Oh and Simon Shaolei Du and Nathan Lambert and Sewon Min and Ranjay Krishna and Yulia Tsvetkov and Hannaneh Hajishirzi and Pang Wei Koh and Luke Zettlemoyer},
  title = {Spurious Rewards: Rethinking Training Signals in RLVR},
  journal = {arXiv preprint arXiv:2506.10947},
  year = {2026},
  eprint = {2506.10947},
  archivePrefix = {arXiv},
  primaryClass = {cs.AI},
  doi = {10.48550/arXiv.2506.10947},
  url = {https://arxiv.org/abs/2506.10947},
}

@article{shumailov2024collapse,
  author = {Shumailov, Ilia and Shumaylov, Zakhar and Zhao, Yiren and Papernot, Nicolas and Anderson, Ross and Gal, Yarin},
  title = {AI models collapse when trained on recursively generated data},
  journal = {Nature},
  volume = {631},
  number = {8022},
  pages = {755--759},
  publisher = {Springer Science and Business Media LLC},
  month = {July},
  year = {2024},
  doi = {10.1038/s41586-024-07566-y},
  url = {http://dx.doi.org/10.1038/s41586-024-07566-y},
  issn = {1476-4687},
}

@article{paperqa2_2024,
  author = {Michael D. Skarlinski and Sam Cox and Jon M. Laurent and James D. Braza and Michaela Hinks and Michael J. Hammerling and Manvitha Ponnapati and Samuel G. Rodriques and Andrew D. White},
  title = {Language agents achieve superhuman synthesis of scientific knowledge},
  journal = {arXiv preprint arXiv:2409.13740},
  year = {2024},
  eprint = {2409.13740},
  archivePrefix = {arXiv},
  primaryClass = {cs.CL},
  doi = {10.48550/arXiv.2409.13740},
  url = {https://arxiv.org/abs/2409.13740},
}

@article{prmbench2025,
  author = {Mingyang Song and Zhaochen Su and Xiaoye Qu and Jiawei Zhou and Yu Cheng},
  title = {PRMBench: A Fine-grained and Challenging Benchmark for Process-Level Reward Models},
  journal = {arXiv preprint arXiv:2501.03124},
  year = {2025},
  eprint = {2501.03124},
  archivePrefix = {arXiv},
  primaryClass = {cs.CL},
  doi = {10.48550/arXiv.2501.03124},
  url = {https://arxiv.org/abs/2501.03124},
}

@article{mathshepherd2023,
  author = {Peiyi Wang and Lei Li and Zhihong Shao and R. X. Xu and Damai Dai and Yifei Li and Deli Chen and Y. Wu and Zhifang Sui},
  title = {Math-Shepherd: Verify and Reinforce LLMs Step-by-step without Human Annotations},
  journal = {arXiv preprint arXiv:2312.08935},
  year = {2024},
  eprint = {2312.08935},
  archivePrefix = {arXiv},
  primaryClass = {cs.AI},
  doi = {10.48550/arXiv.2312.08935},
  url = {https://arxiv.org/abs/2312.08935},
}

@article{openhands2025,
  author = {Xingyao Wang and Boxuan Li and Yufan Song and Frank F. Xu and Xiangru Tang and Mingchen Zhuge and Jiayi Pan and Yueqi Song and Bowen Li and Jaskirat Singh and Hoang H. Tran and Fuqiang Li and Ren Ma and Mingzhang Zheng and Bill Qian and Yanjun Shao and Niklas Muennighoff and Yizhe Zhang and Binyuan Hui and Junyang Lin and Robert Brennan and Hao Peng and Heng Ji and Graham Neubig},
  title = {OpenHands: An Open Platform for AI Software Developers as Generalist Agents},
  journal = {arXiv preprint arXiv:2407.16741},
  year = {2025},
  eprint = {2407.16741},
  archivePrefix = {arXiv},
  primaryClass = {cs.SE},
  doi = {10.48550/arXiv.2407.16741},
  url = {https://arxiv.org/abs/2407.16741},
}

@article{osworld2024,
  author = {Tianbao Xie and Danyang Zhang and Jixuan Chen and Xiaochuan Li and Siheng Zhao and Ruisheng Cao and Toh Jing Hua and Zhoujun Cheng and Dongchan Shin and Fangyu Lei and Yitao Liu and Yiheng Xu and Shuyan Zhou and Silvio Savarese and Caiming Xiong and Victor Zhong and Tao Yu},
  title = {OSWorld: Benchmarking Multimodal Agents for Open-Ended Tasks in Real Computer Environments},
  journal = {arXiv preprint arXiv:2404.07972},
  year = {2024},
  eprint = {2404.07972},
  archivePrefix = {arXiv},
  primaryClass = {cs.AI},
  doi = {10.48550/arXiv.2404.07972},
  url = {https://arxiv.org/abs/2404.07972},
}

@article{kodcode2025,
  author = {Zhangchen Xu and Yang Liu and Yueqin Yin and Mingyuan Zhou and Radha Poovendran},
  title = {KodCode: A Diverse, Challenging, and Verifiable Synthetic Dataset for Coding},
  journal = {arXiv preprint arXiv:2503.02951},
  year = {2025},
  eprint = {2503.02951},
  archivePrefix = {arXiv},
  primaryClass = {cs.LG},
  doi = {10.48550/arXiv.2503.02951},
  url = {https://arxiv.org/abs/2503.02951},
}

@inproceedings{bfcl2024,
  author = {Patil, Shishir G. and Mao, Huanzhi and Cheng-Jie Ji, Charlie and Yan, Fanjia and Suresh, Vishnu and Stoica, Ion and E. Gonzalez, Joseph},
  title = {The Berkeley Function Calling Leaderboard (BFCL): From Tool Use to Agentic Evaluation of Large Language Models},
  booktitle = {Forty-second International Conference on Machine Learning},
  year = {2025},
  url = {https://gorilla.cs.berkeley.edu/blogs/13_bfcl_v3_multi_turn.html},
}

@article{limo2025,
  author = {Yixin Ye and Zhen Huang and Yang Xiao and Ethan Chern and Shijie Xia and Pengfei Liu},
  title = {LIMO: Less is More for Reasoning},
  journal = {arXiv preprint arXiv:2502.03387},
  year = {2025},
  eprint = {2502.03387},
  archivePrefix = {arXiv},
  primaryClass = {cs.CL},
  doi = {10.48550/arXiv.2502.03387},
  url = {https://arxiv.org/abs/2502.03387},
}

@article{dapo2025,
  author = {Qiying Yu and Zheng Zhang and Ruofei Zhu and Yufeng Yuan and Xiaochen Zuo and Yu Yue and Weinan Dai and Tiantian Fan and Gaohong Liu and Lingjun Liu and Xin Liu and Haibin Lin and Zhiqi Lin and Bole Ma and Guangming Sheng and Yuxuan Tong and Chi Zhang and Mofan Zhang and Wang Zhang and Hang Zhu and Jinhua Zhu and Jiaze Chen and Jiangjie Chen and Chengyi Wang and Hongli Yu and Yuxuan Song and Xiangpeng Wei and Hao Zhou and Jingjing Liu and Wei-Ying Ma and Ya-Qin Zhang and Lin Yan and Mu Qiao and Yonghui Wu and Mingxuan Wang},
  title = {DAPO: An Open-Source LLM Reinforcement Learning System at Scale},
  journal = {arXiv preprint arXiv:2503.14476},
  year = {2025},
  eprint = {2503.14476},
  archivePrefix = {arXiv},
  primaryClass = {cs.LG},
  doi = {10.48550/arXiv.2503.14476},
  url = {https://arxiv.org/abs/2503.14476},
}

@article{yue2025rlreasoning,
  author = {Yang Yue and Zhiqi Chen and Rui Lu and Andrew Zhao and Zhaokai Wang and Yang Yue and Shiji Song and Gao Huang},
  title = {Does Reinforcement Learning Really Incentivize Reasoning Capacity in LLMs Beyond the Base Model?},
  journal = {arXiv preprint arXiv:2504.13837},
  year = {2025},
  eprint = {2504.13837},
  archivePrefix = {arXiv},
  primaryClass = {cs.AI},
  doi = {10.48550/arXiv.2504.13837},
  url = {https://arxiv.org/abs/2504.13837},
}

@article{zelikman2022star,
  author = {Eric Zelikman and Yuhuai Wu and Jesse Mu and Noah D. Goodman},
  title = {STaR: Bootstrapping Reasoning With Reasoning},
  journal = {arXiv preprint arXiv:2203.14465},
  year = {2022},
  eprint = {2203.14465},
  archivePrefix = {arXiv},
  primaryClass = {cs.LG},
  doi = {10.48550/arXiv.2203.14465},
  url = {https://arxiv.org/abs/2203.14465},
}

@article{zhai2026passkt,
  author = {Zhiyuan Zhai and Wenjing Yan and Xiaodan Shao and Xin Wang},
  title = {Does RL Expand the Capability Boundary of LLM Agents? A PASS@(k,T) Analysis},
  journal = {arXiv preprint arXiv:2604.14877},
  year = {2026},
  eprint = {2604.14877},
  archivePrefix = {arXiv},
  primaryClass = {cs.LG},
  doi = {10.48550/arXiv.2604.14877},
  url = {https://arxiv.org/abs/2604.14877},
}

@article{qwen25mathprm2025,
  author = {Zhenru Zhang and Chujie Zheng and Yangzhen Wu and Beichen Zhang and Runji Lin and Bowen Yu and Dayiheng Liu and Jingren Zhou and Junyang Lin},
  title = {The Lessons of Developing Process Reward Models in Mathematical Reasoning},
  journal = {arXiv preprint arXiv:2501.07301},
  year = {2025},
  eprint = {2501.07301},
  archivePrefix = {arXiv},
  primaryClass = {cs.CL},
  doi = {10.48550/arXiv.2501.07301},
  url = {https://arxiv.org/abs/2501.07301},
}

@inproceedings{processbench2025,
  author = {Zheng, Chujie and Zhang, Zhenru and Zhang, Beichen and Lin, Runji and Lu, Keming and Yu, Bowen and Liu, Dayiheng and Zhou, Jingren and Lin, Junyang},
  title = {{P}rocess{B}ench: Identifying Process Errors in Mathematical Reasoning},
  booktitle = {Proceedings of the 63rd Annual Meeting of the Association for Computational Linguistics (Volume 1: Long Papers)},
  pages = {1009--1024},
  publisher = {Association for Computational Linguistics},
  address = {Vienna, Austria},
  month = {jul},
  year = {2025},
  eprint = {2412.06559},
  archivePrefix = {arXiv},
  primaryClass = {cs.AI},
  doi = {10.18653/v1/2025.acl-long.50},
  url = {https://aclanthology.org/2025.acl-long.50/},
  note = {arXiv:2412.06559},
  editor = {Che, Wanxiang and Nabende, Joyce and Shutova, Ekaterina and Pilehvar, Mohammad Taher},
  isbn = {979-8-89176-251-0},
}

@article{lanham2023,
  author = {Tamera Lanham and Anna Chen and Ansh Radhakrishnan and Benoit Steiner and Carson Denison and Danny Hernandez and Dustin Li and Esin Durmus and Evan Hubinger and Jackson Kernion and Kamil{\.e} Luko{\v{s}}i{\=u}t{\.e} and Karina Nguyen and Newton Cheng and Nicholas Joseph and Nicholas Schiefer and Oliver Rausch and Robin Larson and Sam McCandlish and Sandipan Kundu and Saurav Kadavath and Shannon Yang and Thomas Henighan and Timothy Maxwell and Timothy Telleen-Lawton and Tristan Hume and Zac Hatfield-Dodds and Jared Kaplan and Jan Brauner and Samuel R. Bowman and Ethan Perez},
  title = {Measuring Faithfulness in Chain-of-Thought Reasoning},
  journal = {arXiv preprint arXiv:2307.13702},
  year = {2023},
  eprint = {2307.13702},
  archivePrefix = {arXiv},
  primaryClass = {cs.AI},
  doi = {10.48550/arXiv.2307.13702},
  url = {https://arxiv.org/abs/2307.13702},
}

@article{proofnet2023,
  author = {Zhangir Azerbayev and Bartosz Piotrowski and Hailey Schoelkopf and Edward W. Ayers and Dragomir Radev and Jeremy Avigad},
  title = {ProofNet: Autoformalizing and Formally Proving Undergraduate-Level Mathematics},
  journal = {arXiv preprint arXiv:2302.12433},
  year = {2023},
  eprint = {2302.12433},
  archivePrefix = {arXiv},
  primaryClass = {cs.CL},
  doi = {10.48550/arXiv.2302.12433},
  url = {https://arxiv.org/abs/2302.12433},
}

@inproceedings{holist2019,
  author = {Kshitij Bansal and Sarah M. Loos and Markus N. Rabe and Christian Szegedy and Stewart Wilcox},
  title = {HOList: An Environment for Machine Learning of Higher Order Logic Theorem Proving},
  booktitle = {Proceedings of the 36th International Conference on Machine Learning, {ICML} 2019, 9-15 June 2019, Long Beach, California, {USA}},
  pages = {454--463},
  publisher = {{PMLR}},
  year = {2019},
  url = {http://proceedings.mlr.press/v97/bansal19a.html},
  bibsource = {dblp computer science bibliography, https://dblp.org},
  editor = {Kamalika Chaudhuri and Ruslan Salakhutdinov},
  series = {Proceedings of Machine Learning Research},
}

@article{bioasq2015,
  author = {George Tsatsaronis and Georgios Balikas and Prodromos Malakasiotis and Ioannis Partalas and Matthias Zschunke and Michael R. Alvers and Dirk Weissenborn and Anastasia Krithara and Sergios Petridis and Dimitris Polychronopoulos and Yannis Almirantis and John Pavlopoulos and Nicolas Baskiotis and Patrick Gallinari and Thierry Arti{\`{e}}res and Axel{-}Cyrille Ngonga Ngomo and Norman Heino and {\'{E}}ric Gaussier and Liliana Barrio{-}Alvers and Michael Schroeder and Ion Androutsopoulos and Georgios Paliouras},
  title = {An overview of the {BIOASQ} large-scale biomedical semantic indexing and question answering competition},
  journal = {{BMC} Bioinform.},
  volume = {16},
  pages = {138:1--138:28},
  year = {2015},
  doi = {10.1186/S12859-015-0564-6},
  url = {https://doi.org/10.1186/s12859-015-0564-6},
  bibsource = {dblp computer science bibliography, https://dblp.org},
}

@article{agentcompany2025,
  author = {Frank F. Xu and Yufan Song and Boxuan Li and Yuxuan Tang and Kritanjali Jain and Mengxue Bao and Zora Z. Wang and Xuhui Zhou and Zhitong Guo and Murong Cao and Mingyang Yang and Hao Yang Lu and Amaad Martin and Zhe Su and Leander Maben and Raj Mehta and Wayne Chi and Lawrence Jang and Yiqing Xie and Shuyan Zhou and Graham Neubig},
  title = {TheAgentCompany: Benchmarking LLM Agents on Consequential Real World Tasks},
  journal = {arXiv preprint arXiv:2412.14161},
  year = {2025},
  eprint = {2412.14161},
  archivePrefix = {arXiv},
  primaryClass = {cs.CL},
  doi = {10.48550/arXiv.2412.14161},
  url = {https://arxiv.org/abs/2412.14161},
}

@article{chen2021humaneval,
  author = {Mark Chen and Jerry Tworek and Heewoo Jun and Qiming Yuan and Henrique Ponde de Oliveira Pinto and Jared Kaplan and Harri Edwards and Yuri Burda and Nicholas Joseph and Greg Brockman and Alex Ray and Raul Puri and Gretchen Krueger and Michael Petrov and Heidy Khlaaf and Girish Sastry and Pamela Mishkin and Brooke Chan and Scott Gray and Nick Ryder and Mikhail Pavlov and Alethea Power and Lukasz Kaiser and Mohammad Bavarian and Clemens Winter and Philippe Tillet and Felipe Petroski Such and Dave Cummings and Matthias Plappert and Fotios Chantzis and Elizabeth Barnes and Ariel Herbert-Voss and William Hebgen Guss and Alex Nichol and Alex Paino and Nikolas Tezak and Jie Tang and Igor Babuschkin and Suchir Balaji and Shantanu Jain and William Saunders and Christopher Hesse and Andrew N. Carr and Jan Leike and Josh Achiam and Vedant Misra and Evan Morikawa and Alec Radford and Matthew Knight and Miles Brundage and Mira Murati and Katie Mayer and Peter Welinder and Bob McGrew and Dario Amodei and Sam McCandlish and Ilya Sutskever and Wojciech Zaremba},
  title = {Evaluating Large Language Models Trained on Code},
  journal = {arXiv preprint arXiv:2107.03374},
  year = {2021},
  eprint = {2107.03374},
  archivePrefix = {arXiv},
  primaryClass = {cs.LG},
  doi = {10.48550/arXiv.2107.03374},
  url = {https://arxiv.org/abs/2107.03374},
}

@article{finqa2021,
  author = {Zhiyu Chen and Wenhu Chen and Charese Smiley and Sameena Shah and Iana Borova and Dylan Langdon and Reema Moussa and Matt Beane and Ting-Hao Huang and Bryan Routledge and William Yang Wang},
  title = {FinQA: A Dataset of Numerical Reasoning over Financial Data},
  journal = {arXiv preprint arXiv:2109.00122},
  year = {2022},
  eprint = {2109.00122},
  archivePrefix = {arXiv},
  primaryClass = {cs.CL},
  doi = {10.48550/arXiv.2109.00122},
  url = {https://arxiv.org/abs/2109.00122},
}

@article{livecodebench2024,
  author = {Naman Jain and King Han and Alex Gu and Wen-Ding Li and Fanjia Yan and Tianjun Zhang and Sida Wang and Armando Solar-Lezama and Koushik Sen and Ion Stoica},
  title = {LiveCodeBench: Holistic and Contamination Free Evaluation of Large Language Models for Code},
  journal = {arXiv preprint arXiv:2403.07974},
  year = {2024},
  eprint = {2403.07974},
  archivePrefix = {arXiv},
  primaryClass = {cs.SE},
  doi = {10.48550/arXiv.2403.07974},
  url = {https://arxiv.org/abs/2403.07974},
}

@article{cobbe2021gsm8k,
  author = {Karl Cobbe and Vineet Kosaraju and Mohammad Bavarian and Mark Chen and Heewoo Jun and Lukasz Kaiser and Matthias Plappert and Jerry Tworek and Jacob Hilton and Reiichiro Nakano and Christopher Hesse and John Schulman},
  title = {Training Verifiers to Solve Math Word Problems},
  journal = {arXiv preprint arXiv:2110.14168},
  year = {2021},
  eprint = {2110.14168},
  archivePrefix = {arXiv},
  primaryClass = {cs.LG},
  doi = {10.48550/arXiv.2110.14168},
  url = {https://arxiv.org/abs/2110.14168},
}

@article{ultrafeedback2023,
  author = {Ganqu Cui and Lifan Yuan and Ning Ding and Guanming Yao and Bingxiang He and Wei Zhu and Yuan Ni and Guotong Xie and Ruobing Xie and Yankai Lin and Zhiyuan Liu and Maosong Sun},
  title = {UltraFeedback: Boosting Language Models with Scaled AI Feedback},
  journal = {arXiv preprint arXiv:2310.01377},
  year = {2024},
  eprint = {2310.01377},
  archivePrefix = {arXiv},
  primaryClass = {cs.CL},
  doi = {10.48550/arXiv.2310.01377},
  url = {https://arxiv.org/abs/2310.01377},
}

@article{prime2025,
  author = {Ganqu Cui and Lifan Yuan and Zefan Wang and Hanbin Wang and Yuchen Zhang and Jiacheng Chen and Wendi Li and Bingxiang He and Yuchen Fan and Tianyu Yu and Qixin Xu and Weize Chen and Jiarui Yuan and Huayu Chen and Kaiyan Zhang and Xingtai Lv and Shuo Wang and Yuan Yao and Xu Han and Hao Peng and Yu Cheng and Zhiyuan Liu and Maosong Sun and Bowen Zhou and Ning Ding},
  title = {Process Reinforcement through Implicit Rewards},
  journal = {arXiv preprint arXiv:2502.01456},
  year = {2025},
  eprint = {2502.01456},
  archivePrefix = {arXiv},
  primaryClass = {cs.LG},
  doi = {10.48550/arXiv.2502.01456},
  url = {https://arxiv.org/abs/2502.01456},
}

@article{deepseekprover2024,
  author = {Huajian Xin and Daya Guo and Zhihong Shao and Zhizhou Ren and Qihao Zhu and Bo Liu and Chong Ruan and Wenda Li and Xiaodan Liang},
  title = {DeepSeek-Prover: Advancing Theorem Proving in LLMs through Large-Scale Synthetic Data},
  journal = {arXiv preprint arXiv:2405.14333},
  year = {2024},
  eprint = {2405.14333},
  archivePrefix = {arXiv},
  primaryClass = {cs.AI},
  doi = {10.48550/arXiv.2405.14333},
  url = {https://arxiv.org/abs/2405.14333},
}

@article{deepseekprover15_2024,
  author = {Huajian Xin and Z. Z. Ren and Junxiao Song and Zhihong Shao and Wanjia Zhao and Haocheng Wang and Bo Liu and Liyue Zhang and Xuan Lu and Qiushi Du and Wenjun Gao and Qihao Zhu and Dejian Yang and Zhibin Gou and Z. F. Wu and Fuli Luo and Chong Ruan},
  title = {DeepSeek-Prover-V1.5: Harnessing Proof Assistant Feedback for Reinforcement Learning and Monte-Carlo Tree Search},
  journal = {arXiv preprint arXiv:2408.08152},
  year = {2024},
  eprint = {2408.08152},
  archivePrefix = {arXiv},
  primaryClass = {cs.CL},
  doi = {10.48550/arXiv.2408.08152},
  url = {https://arxiv.org/abs/2408.08152},
}

@article{mind2web2023,
  author = {Xiang Deng and Yu Gu and Boyuan Zheng and Shijie Chen and Samuel Stevens and Boshi Wang and Huan Sun and Yu Su},
  title = {Mind2Web: Towards a Generalist Agent for the Web},
  journal = {arXiv preprint arXiv:2306.06070},
  year = {2023},
  eprint = {2306.06070},
  archivePrefix = {arXiv},
  primaryClass = {cs.CL},
  doi = {10.48550/arXiv.2306.06070},
  url = {https://arxiv.org/abs/2306.06070},
}

@article{directoveropt2024,
  author = {Rafael Rafailov and Yaswanth Chittepu and Ryan Park and Harshit Sikchi and Joey Hejna and Bradley Knox and Chelsea Finn and Scott Niekum},
  title = {Scaling Laws for Reward Model Overoptimization in Direct Alignment Algorithms},
  journal = {arXiv preprint arXiv:2406.02900},
  year = {2024},
  eprint = {2406.02900},
  archivePrefix = {arXiv},
  primaryClass = {cs.LG},
  doi = {10.48550/arXiv.2406.02900},
  url = {https://arxiv.org/abs/2406.02900},
}

@article{androidworld2025,
  author = {Christopher Rawles and Sarah Clinckemaillie and Yifan Chang and Jonathan Waltz and Gabrielle Lau and Marybeth Fair and Alice Li and William Bishop and Wei Li and Folawiyo Campbell-Ajala and Daniel Toyama and Robert Berry and Divya Tyamagundlu and Timothy Lillicrap and Oriana Riva},
  title = {AndroidWorld: A Dynamic Benchmarking Environment for Autonomous Agents},
  journal = {arXiv preprint arXiv:2405.14573},
  year = {2025},
  eprint = {2405.14573},
  archivePrefix = {arXiv},
  primaryClass = {cs.AI},
  doi = {10.48550/arXiv.2405.14573},
  url = {https://arxiv.org/abs/2405.14573},
}

@article{openthoughts2025,
  author = {Etash Guha and Ryan Marten and Sedrick Keh and Negin Raoof and Georgios Smyrnis and Hritik Bansal and Marianna Nezhurina and Jean Mercat and Trung Vu and Zayne Sprague and Ashima Suvarna and Benjamin Feuer and Liangyu Chen and Zaid Khan and Eric Frankel and Sachin Grover and Caroline Choi and Niklas Muennighoff and Shiye Su and Wanjia Zhao and John Yang and Shreyas Pimpalgaonkar and Kartik Sharma and Charlie Cheng-Jie Ji and Yichuan Deng and Sarah Pratt and Vivek Ramanujan and Jon Saad-Falcon and Jeffrey Li and Achal Dave and Alon Albalak and Kushal Arora and Blake Wulfe and Chinmay Hegde and Greg Durrett and Sewoong Oh and Mohit Bansal and Saadia Gabriel and Aditya Grover and Kai-Wei Chang and Vaishaal Shankar and Aaron Gokaslan and Mike A. Merrill and Tatsunori Hashimoto and Yejin Choi and Jenia Jitsev and Reinhard Heckel and Maheswaran Sathiamoorthy and Alexandros G. Dimakis and Ludwig Schmidt},
  title = {OpenThoughts: Data Recipes for Reasoning Models},
  journal = {arXiv preprint arXiv:2506.04178},
  year = {2025},
  eprint = {2506.04178},
  archivePrefix = {arXiv},
  primaryClass = {cs.LG},
  doi = {10.48550/arXiv.2506.04178},
  url = {https://arxiv.org/abs/2506.04178},
}

@article{hendrycks2021math,
  author = {Dan Hendrycks and Collin Burns and Saurav Kadavath and Akul Arora and Steven Basart and Eric Tang and Dawn Song and Jacob Steinhardt},
  title = {Measuring Mathematical Problem Solving With the MATH Dataset},
  journal = {arXiv preprint arXiv:2103.03874},
  year = {2021},
  eprint = {2103.03874},
  archivePrefix = {arXiv},
  primaryClass = {cs.LG},
  doi = {10.48550/arXiv.2103.03874},
  url = {https://arxiv.org/abs/2103.03874},
}

@article{hendrycks2021apps,
  author = {Dan Hendrycks and Steven Basart and Saurav Kadavath and Mantas Mazeika and Akul Arora and Ethan Guo and Collin Burns and Samir Puranik and Horace He and Dawn Song and Jacob Steinhardt},
  title = {Measuring Coding Challenge Competence With APPS},
  journal = {arXiv preprint arXiv:2105.09938},
  year = {2021},
  eprint = {2105.09938},
  archivePrefix = {arXiv},
  primaryClass = {cs.SE},
  doi = {10.48550/arXiv.2105.09938},
  url = {https://arxiv.org/abs/2105.09938},
}

@misc{swebenchverified2024,
  author = {Chowdhury, Neil and Aung, James and Shern, Chan Jun and Jaffe, Oliver and Sherburn, Dane and Starace, Giulio and Mays, Evan and Dias, Rachel and Aljubeh, Marwan and Glaese, Mia and Jimenez, Carlos E. and Yang, John and Ho, Leyton and Patwardhan, Tejal and Liu, Kevin and Madry, Aleksander},
  title = {Introducing {SWE}-bench Verified},
  year = {2024},
  url = {https://openai.com/index/introducing-swe-bench-verified/},
  note = {OpenAI blog post},
}

@inproceedings{qasper2021,
  author = {Dasigi, Pradeep and Lo, Kyle and Beltagy, Iz and Cohan, Arman and Smith, Noah A. and Gardner, Matt},
  title = {A Dataset of Information-Seeking Questions and Answers Anchored in Research Papers},
  booktitle = {Proceedings of the 2021 Conference of the North American Chapter of the Association for Computational Linguistics: Human Language Technologies},
  pages = {4599--4610},
  publisher = {Association for Computational Linguistics},
  address = {Online},
  month = {jun},
  year = {2021},
  doi = {10.18653/v1/2021.naacl-main.365},
  url = {https://aclanthology.org/2021.naacl-main.365/},
  editor = {Toutanova, Kristina and Rumshisky, Anna and Zettlemoyer, Luke and Hakkani-Tur, Dilek and Beltagy, Iz and Bethard, Steven and Cotterell, Ryan and Chakraborty, Tanmoy and Zhou, Yichao},
}

@article{prometheus2_2024,
  author = {Seungone Kim and Juyoung Suk and Shayne Longpre and Bill Yuchen Lin and Jamin Shin and Sean Welleck and Graham Neubig and Moontae Lee and Kyungjae Lee and Minjoon Seo},
  title = {Prometheus 2: An Open Source Language Model Specialized in Evaluating Other Language Models},
  journal = {arXiv preprint arXiv:2405.01535},
  year = {2024},
  eprint = {2405.01535},
  archivePrefix = {arXiv},
  primaryClass = {cs.CL},
  doi = {10.48550/arXiv.2405.01535},
  url = {https://arxiv.org/abs/2405.01535},
}

@article{visualwebarena2024,
  author = {Jing Yu Koh and Robert Lo and Lawrence Jang and Vikram Duvvur and Ming Chong Lim and Po-Yu Huang and Graham Neubig and Shuyan Zhou and Ruslan Salakhutdinov and Daniel Fried},
  title = {VisualWebArena: Evaluating Multimodal Agents on Realistic Visual Web Tasks},
  journal = {arXiv preprint arXiv:2401.13649},
  year = {2024},
  eprint = {2401.13649},
  archivePrefix = {arXiv},
  primaryClass = {cs.LG},
  doi = {10.48550/arXiv.2401.13649},
  url = {https://arxiv.org/abs/2401.13649},
}

@article{cuad2021,
  author = {Dan Hendrycks and Collin Burns and Anya Chen and Spencer Ball},
  title = {CUAD: An Expert-Annotated NLP Dataset for Legal Contract Review},
  journal = {arXiv preprint arXiv:2103.06268},
  year = {2021},
  eprint = {2103.06268},
  archivePrefix = {arXiv},
  primaryClass = {cs.CL},
  doi = {10.48550/arXiv.2103.06268},
  url = {https://arxiv.org/abs/2103.06268},
}

@article{toolsandbox2024,
  author = {Jiarui Lu and Thomas Holleis and Yizhe Zhang and Bernhard Aumayer and Feng Nan and Felix Bai and Shuang Ma and Shen Ma and Mengyu Li and Guoli Yin and Zirui Wang and Ruoming Pang},
  title = {ToolSandbox: A Stateful, Conversational, Interactive Evaluation Benchmark for LLM Tool Use Capabilities},
  journal = {arXiv preprint arXiv:2408.04682},
  year = {2025},
  eprint = {2408.04682},
  archivePrefix = {arXiv},
  primaryClass = {cs.CL},
  doi = {10.48550/arXiv.2408.04682},
  url = {https://arxiv.org/abs/2408.04682},
}


\appendix

\setcounter{figure}{0}
\setcounter{table}{0}
\renewcommand{\thefigure}{A\arabic{figure}}
\renewcommand{\thetable}{A\arabic{table}}

\section{Supplementary Figures and Audit Notes}
\label{app:supplementary}

This appendix provides supplementary material for the primer. The main text is self-contained; the figures and tables below provide expanded schematics, audit fields, source-placement rules, and a brief AI-assistance disclosure. The technical material follows the order of the main text: verification contracts, construction layers, scaling attribution, agent-trajectory audit fields, and source placement.

\subsection{Scope, Methodology, and Inclusion Criteria}
\label{app:method}

This primer is a question-driven synthesis rather than a formal meta-analysis. 
We focus on public work that makes at least one component of reasoning post-training data explicit: the data object, verifier, trace source, reward channel, environment, scaling rule, or release metadata. 
The goal is not to exhaustively enumerate all reasoning datasets, but to identify the recurring feedback interfaces that determine whether a reported gain can be attributed.

We included work when it satisfied at least one of four criteria. 
First, the work releases or describes reasoning-oriented post-training data, including math, code, theorem proving, tool use, agent trajectories, medical or safety reasoning, and rubric-based evaluation. 
Second, it exposes a verifier, reward model, process label, environment predicate, judge, or feedback rule that determines what counts as success. 
Third, it describes a construction recipe, such as prompt sourcing, trace generation, distillation, filtering, self-play, or verifier refresh. 
Fourth, it contributes scaling evidence for reasoning post-training, including data reuse, test-time compute, RLVR, verifier scaling, or inference-budget effects.

We excluded work that only reports leaderboard performance without describing the data object, feedback interface, construction recipe, or evaluation budget. 
We also treat technical reports, arXiv preprints, benchmark papers, and model cards differently from peer-reviewed empirical papers: non-peer-reviewed sources are used primarily for documenting public recipes and release practices, not for settling causal claims. 
This distinction is important because the paper's central question is not simply whether a model improves, but which data object, verifier, scaffold, or inference budget carries the improvement.

\begin{table}[t]
\centering
\small
\setlength{\tabcolsep}{4pt}
\renewcommand{\arraystretch}{1.10}
\begin{tabularx}{\columnwidth}{@{}
>{\raggedright\arraybackslash}p{0.25\columnwidth}
>{\raggedright\arraybackslash}X
>{\raggedright\arraybackslash}p{0.25\columnwidth}@{}}
\toprule
\textbf{Source type} & \textbf{Included when it exposes} & \textbf{Used for} \\
\midrule
Model report 
& post-training recipe, optimizer, verifier, or release budget 
& construction and scaling context \\

Dataset / benchmark 
& task source, split, verifier, contamination audit, or evaluation rule 
& data-object taxonomy \\

Verifier / reward work 
& checker, PRM, judge, rubric, or failure mode 
& quality and audit fields \\

Agent / environment work 
& state, action, observation, terminal predicate, or replay log 
& environmental verification \\

Scaling study 
& compute, reuse, inference budget, or asymptote-efficiency claim 
& scaling attribution \\
\bottomrule
\end{tabularx}
\caption{\textbf{Inclusion lens for the primer.}
Sources are included when they expose a reasoning-data object, feedback interface, construction recipe, or scaling surface, not merely because they report a higher benchmark score.}
\label{tab:appendix-scope}
\vspace{-0.35cm}

\end{table}

\subsection{Supplementary Source Coverage}
\label{app:source-coverage}

The main text cites sources at the point where they support an argument. 
To make the survey boundary explicit, Table~\ref{tab:source-coverage-clusters} lists additional source clusters used as coverage checks for the primer. 
These sources are not treated as interchangeable evidence: benchmarks identify task and evaluation surfaces, data releases expose trainable objects, verifier papers expose reward channels, and agent environments expose stateful interaction fields. 
The purpose of this table is therefore bibliographic coverage rather than causal aggregation.

\begin{table*}[t]
\centering
\scriptsize
\setlength{\tabcolsep}{4pt}
\renewcommand{\arraystretch}{1.08}
\begin{tabularx}{\textwidth}{@{}
>{\raggedright\arraybackslash}p{0.18\textwidth}
>{\raggedright\arraybackslash}p{0.31\textwidth}
>{\raggedright\arraybackslash}X@{}}
\toprule
\textbf{Coverage cluster} & \textbf{What it contributes to the primer} & \textbf{Representative sources} \\
\midrule
General reasoning and live evaluations
& contamination-aware or high-difficulty evaluation surfaces for testing whether reported reasoning gains are robust beyond static benchmark reuse
& \citep{hle2025,gpqa2023,frontiermath2024,livebench2024} \\

Math and code benchmarks
& canonical answer- or execution-checkable tasks that make programmatic verification reusable for post-training and evaluation
& \citep{cobbe2021gsm8k,hendrycks2021math,chen2021humaneval,hendrycks2021apps,bigcodebench2024,livecodebench2024} \\

Formal proving environments
& proof-state, theorem-proving, and proof-assistant feedback surfaces where verifier and substrate can collapse into the same object
& \citep{minif2f2021,yang2023leandojo,proofnet2023,holist2019,deepseekprover2024,deepseekprover15_2024} \\

Instruction, preference, and alignment data
& earlier post-training and preference-learning precedents that clarify how demonstrations, comparisons, reward models, and feedback traces become trainable data objects
& \citep{selfinstruct2022,flan2021,t0_2021,instructgpt2022,ultrafeedback2023,hhrlhf2022,dpo2023} \\

Tool use and agent environments
& action schemas, tool calls, web or software environments, and stateful interaction protocols needed to analyze environmental verification
& \citep{react2023,toolformer2023,apibank2023,gorilla2023,apigen2024,bfcl2024,toolsandbox2024,browsergym2024,webarena2023,workarena2024,androidworld2025,agentcompany2025} \\

Scientific and biomedical reasoning
& evidence-grounded, retrieval-heavy, and expert-sensitive domains where correctness is often evidential rather than a simple answer string
& \citep{qasper2021,scifact2020,paperqa2_2024,medpalm2023,pubmedqa2019,bioasq2015} \\

Financial and legal reasoning
& domain-specific reasoning settings where tabular evidence, contracts, holdings, and expert annotation make provenance and rubric fields central
& \citep{financebench2023,finqa2021,tatqa2021,legalbench2023,casehold2021,cuad2021,contractnli2021} \\

Safety, judges, and reward-model audits
& reward-channel and judge-failure evidence for why verifier cards, attack tests, and refresh logs should be part of reasoning-data releases
& \citep{truthfulqa2022,harmbench2024,wildguard2024,xstest2024,rewardbench2024,judgelm2023,prometheus2_2024,rewardoveropt2022,directoveropt2024} \\
\bottomrule
\end{tabularx}
\caption{\textbf{Supplementary source-coverage clusters.}
These clusters expand the bibliographic boundary of the primer beyond the sources discussed in detail in the main text. 
Each cluster is used as a coverage check for a different part of the attribution framework: verification contract, feedback interface, construction recipe, or scaling surface.}
\label{tab:source-coverage-clusters}
\vspace{-0.35cm}
\end{table*}

\subsection{Supplementary Taxonomy View}
\label{app:taxonomy-diagnostics}

\begin{figure}[!t]
    \centering
    \includegraphics[width=\columnwidth]{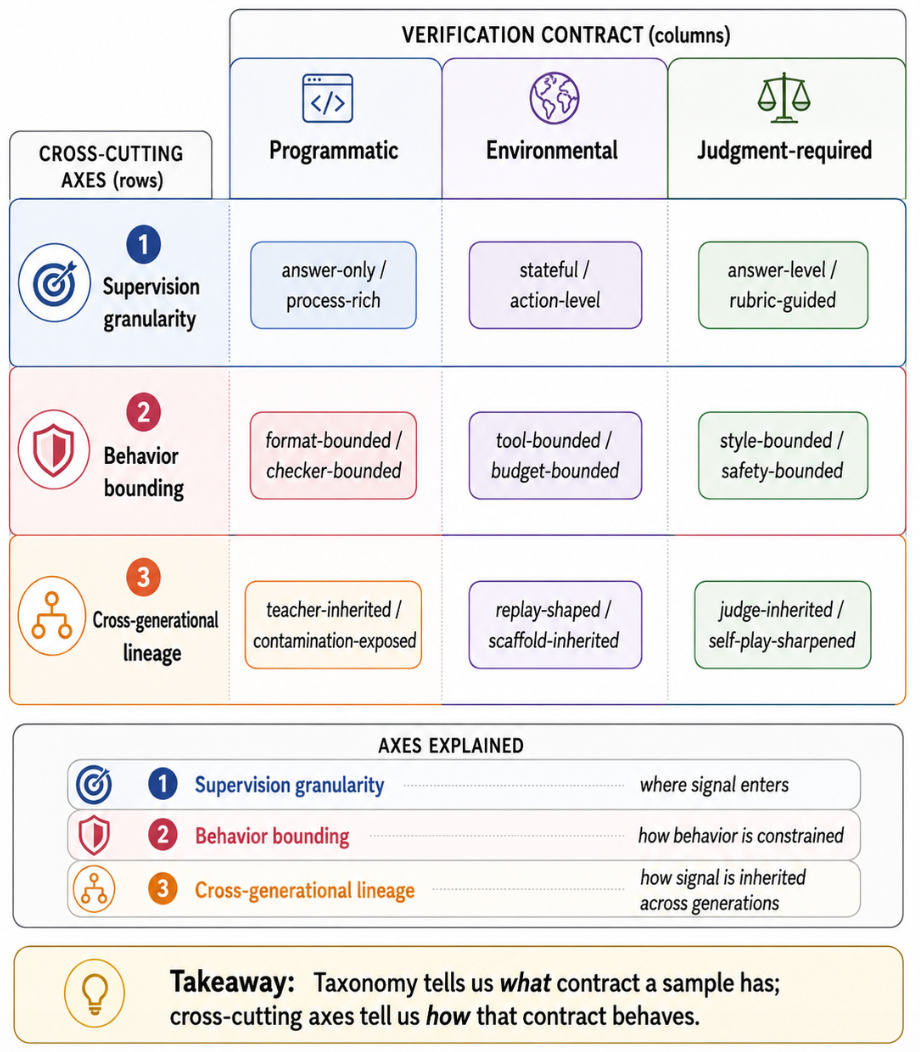}
    \caption{\textbf{Cross-cutting axes over verifier contracts.}
    Verification contracts describe what can be checked, whereas granularity, behaviour bounds, and lineage describe how the signal behaves once it enters training. They should therefore be reported orthogonally to programmatic, environmental, and judgment-required contracts.}
    \label{fig:cross-cutting-axes}
    \vspace{-0.25cm}
\end{figure}

Section~\ref{sec:what-data-need} organizes reasoning post-training data by verification contract. 
Figure~\ref{fig:cross-cutting-axes} expands that taxonomy by showing why contract labels alone are not sufficient. 
A sample can be programmatically verified but still answer-only; environmentally verified but scaffold-inherited; or judgment-required but shaped by self-play, rubric design, or teacher lineage. 
The three cross-cutting axes---supervision granularity, behaviour bounds, and cross-generational lineage---therefore need to be reported orthogonally to the contract itself.

The diagnostic implication is that a data release should not only name the domain or verifier family. 
It should specify where supervision enters the trajectory, which behaviours are allowed or bounded, and which generator, teacher, filter, split, or decontamination procedure produced the sample. 
Without these fields, two superficially similar datasets may instantiate different learning problems.

\subsection{Supplementary Construction Schematics}
\label{app:construction-supplement}

Section~\ref{sec:how-built} compresses the construction discussion into Table~\ref{tab:construction-fields}. 
The figures below provide schematic views of the same construction problem. 
They should be read as explanatory diagrams rather than additional claims: each one isolates a different place where gain attribution can become ambiguous.

\begin{figure}[!t]
    \centering
    \includegraphics[width=0.96\columnwidth]{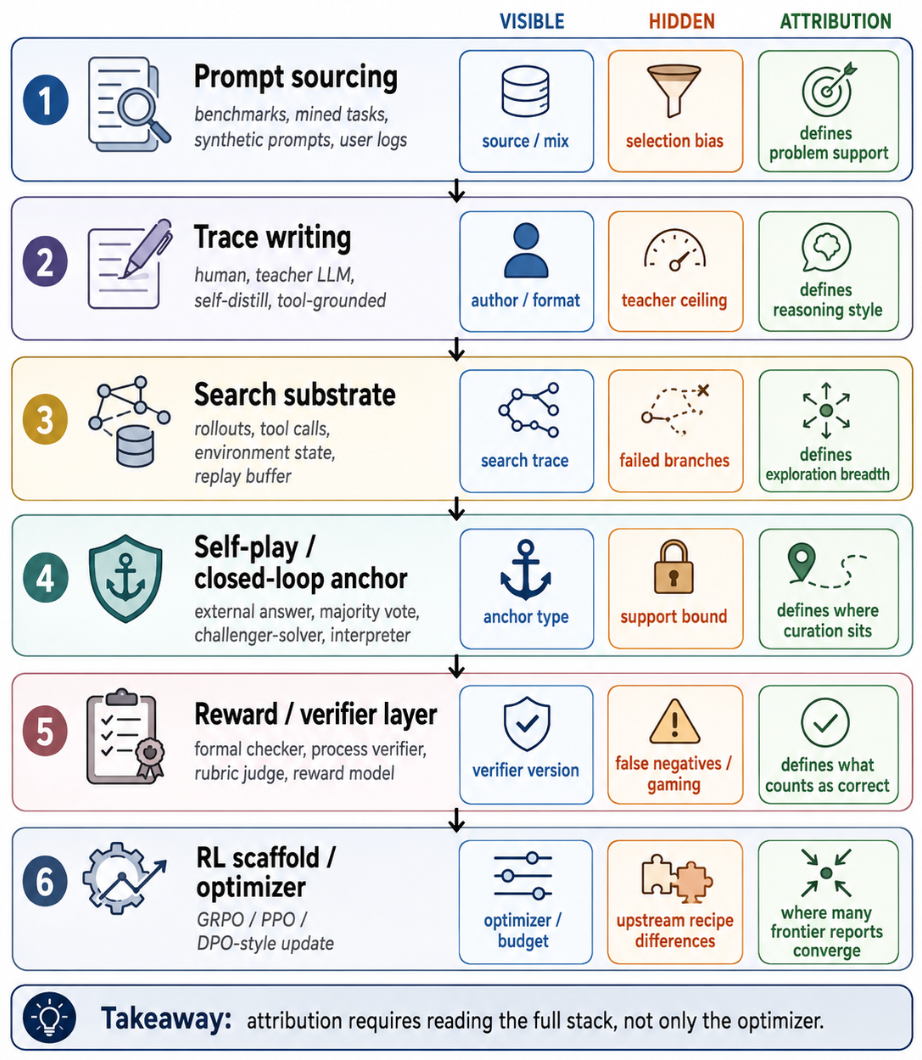}
    \caption{\textbf{The reasoning-data construction stack.}
    Reasoning-data recipes accumulate hidden variables before the downstream optimizer is applied. Prompt sourcing determines problem support; trace writing determines inherited reasoning style; search substrates preserve or erase failed branches; anchors determine where curation re-enters; and verifiers determine what counts as success.}
    \label{fig:construction-stack}
    \vspace{-0.25cm}
\end{figure}

Figure~\ref{fig:construction-stack} gives the full construction stack. 
Its main point is that frontier systems can converge at the visible RL scaffold while diverging upstream in prompt sourcing, trace writing, search substrate, self-play anchors, and verifier design. 
This is why optimizer comparisons alone cannot determine where the gain came from.

\begin{figure}[!t]
    \centering
    \includegraphics[width=0.96\columnwidth]{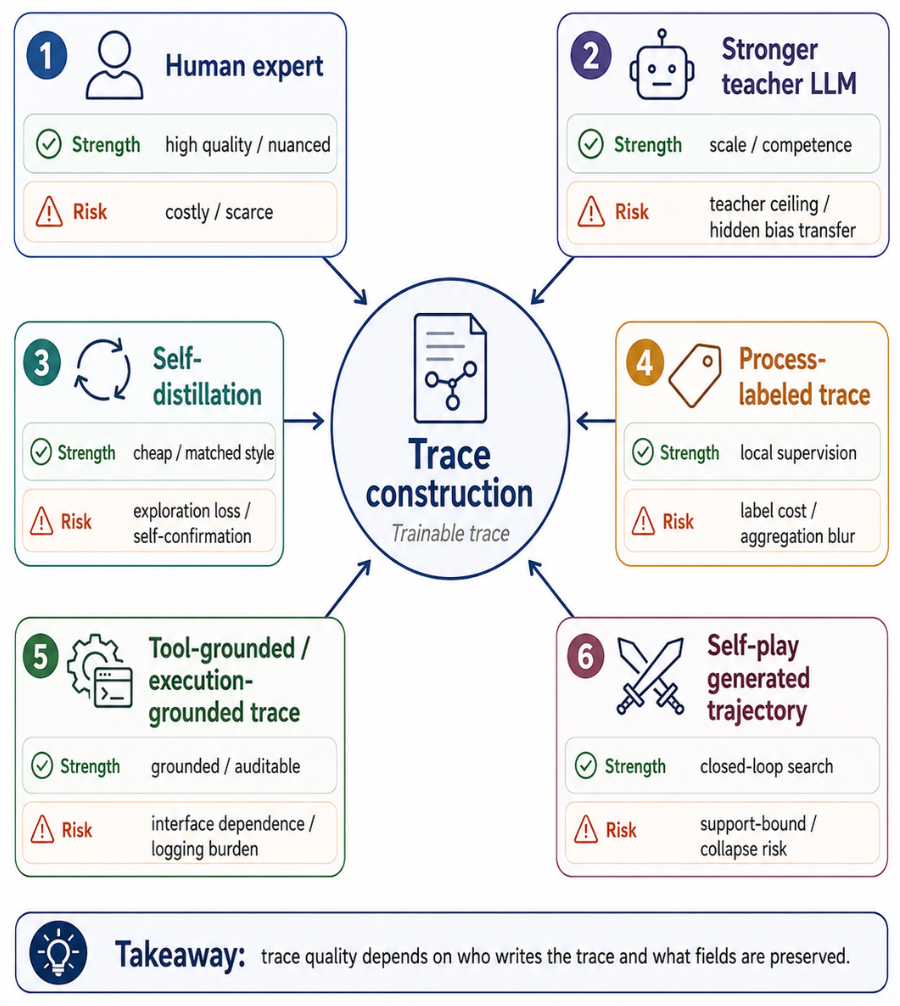}
    \caption{\textbf{Who writes the trace?}
    Trace authorship changes both the ceiling and the attribution risk. 
    Human experts, teacher models, self-distillation, process labels, tool grounding, and self-play trajectories differ in cost, scalability, provenance, and the degree to which they expose the mechanism rather than only the answer.}
    \label{fig:trace-authorship}
    \vspace{-0.25cm}
\end{figure}

Figure~\ref{fig:trace-authorship} isolates trace authorship. 
The same final answer can be paired with very different process fields depending on whether the trace is written by a human expert, teacher model, self-distillation loop, process-labeling procedure, tool-grounded execution, or self-play trajectory. 
This matters because traces do not merely supply intermediate text; they can transmit formatting conventions, decomposition style, stopping behaviour, uncertainty expression, and tool-use habits.

\begin{figure}[!t]
    \centering
    \includegraphics[width=0.96\columnwidth]{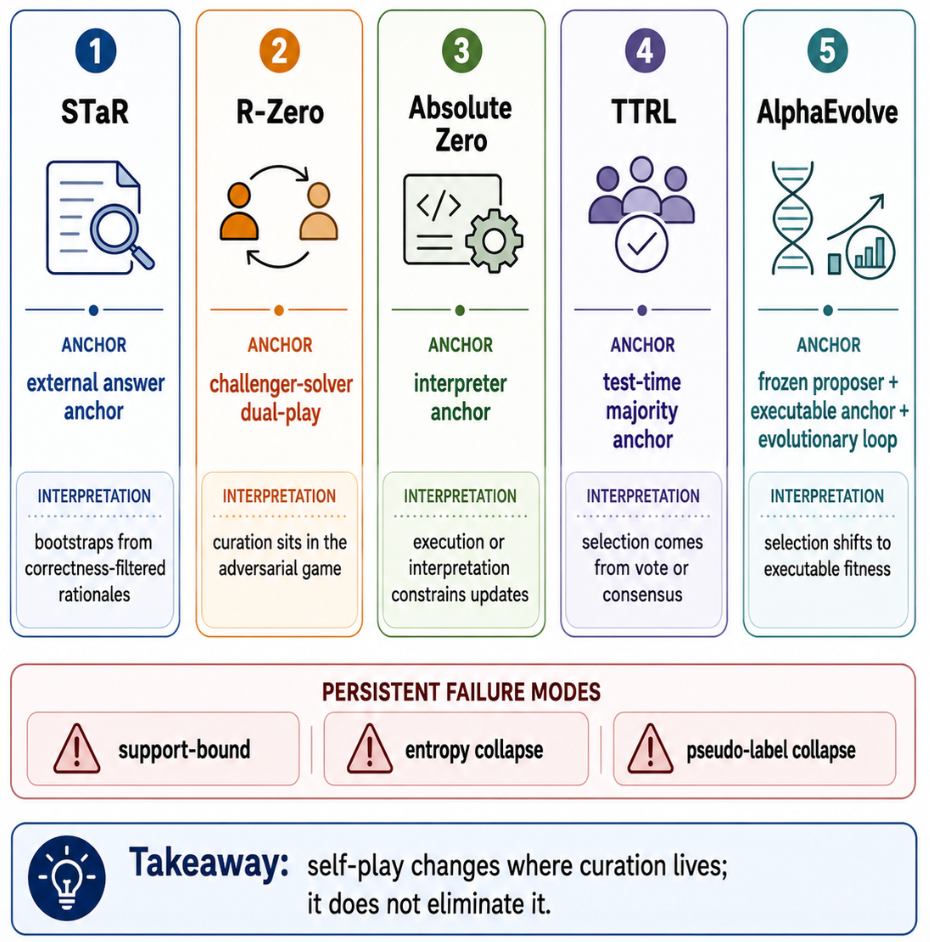}
    \caption{\textbf{Self-play relocates curation.}
    STaR, R-Zero, Absolute Zero, TTRL, and AlphaEvolve-style systems differ less in whether they are ``self-generated'' than in the anchor that converts rollouts into feedback. 
    The anchor defines which generated behaviours become trainable and which are discarded.}
    \label{fig:self-play-anchors}
    \vspace{-0.25cm}
\end{figure}

Figure~\ref{fig:self-play-anchors} expands the self-play discussion. 
Self-play is often described as reducing human supervision, but it does not remove curation. 
Instead, it moves curation to the anchor that turns rollouts into trainable signals: an external answer, an interpreter, a verifier, a majority vote, an archive, or a role split between agents. 
For this reason, self-play releases should report not only the generated tasks, but also the anchor, selection rule, verifier, replay policy, and failure cases.

\subsection{Supplementary Scaling Views}
\label{app:scaling-supplement}

\begin{figure}[!t]
    \centering
    \includegraphics[width=0.98\columnwidth]{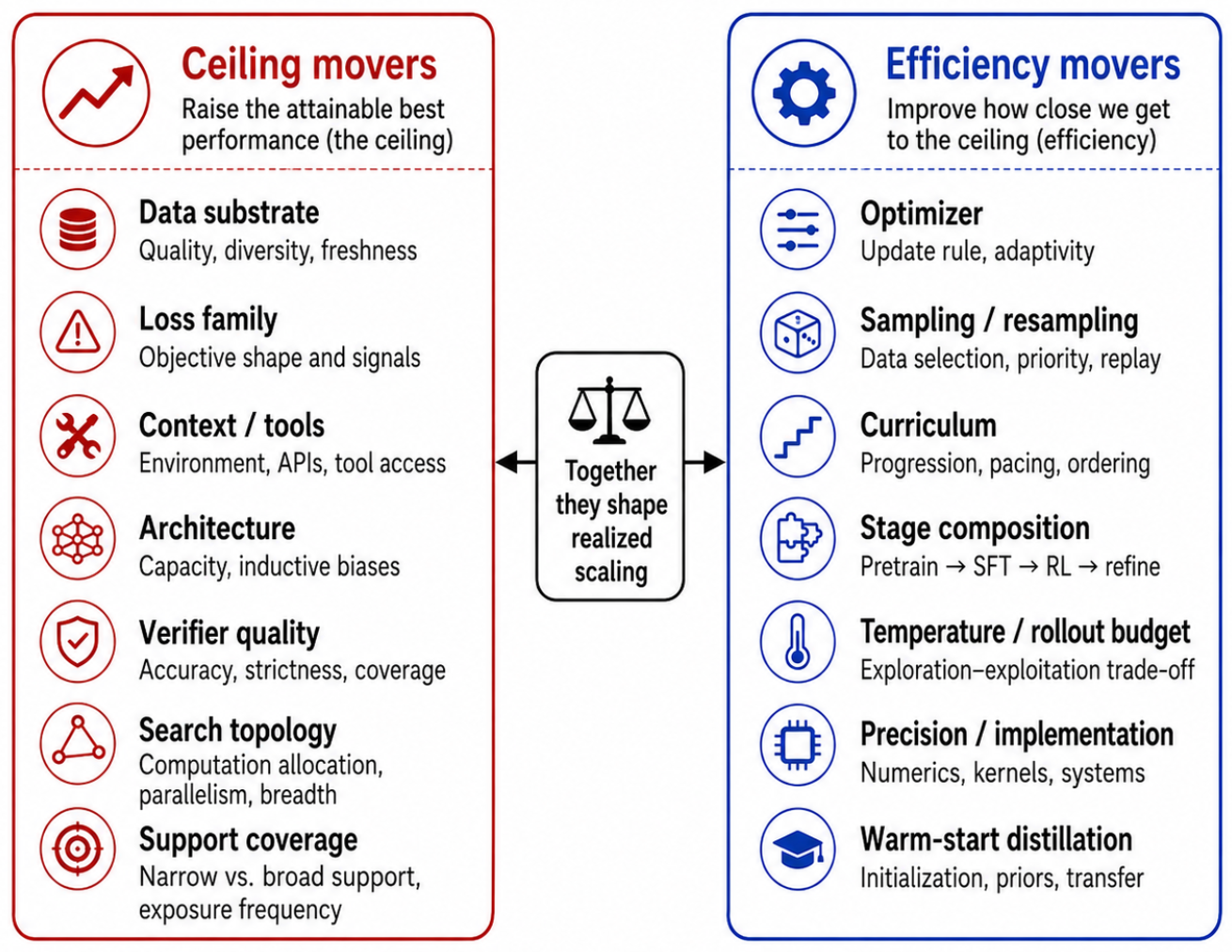}
    \caption{\textbf{What moves the ceiling, what moves efficiency.}
    Ceiling movers change what is reachable under a data substrate, verifier, environment, or context; efficiency movers change how quickly the policy approaches that frontier through optimizer design, sampling, curriculum, precision, or stage composition.}
    \label{fig:ceiling-efficiency-movers}
    \vspace{-0.25cm}
\end{figure}

Section~\ref{sec:scaling} uses the asymptote--efficiency lens to separate what a data substrate makes reachable from how efficiently an optimizer approaches that frontier. 
Figure~\ref{fig:ceiling-efficiency-movers} expands this separation by listing typical ceiling movers and efficiency movers. 
The distinction is not absolute: some interventions, especially verifier refresh or environment redesign, may affect both. 
The figure is therefore a reading device for scaling claims rather than a universal scaling law.

\begin{figure}[!t]
    \centering
    \includegraphics[width=0.98\columnwidth]{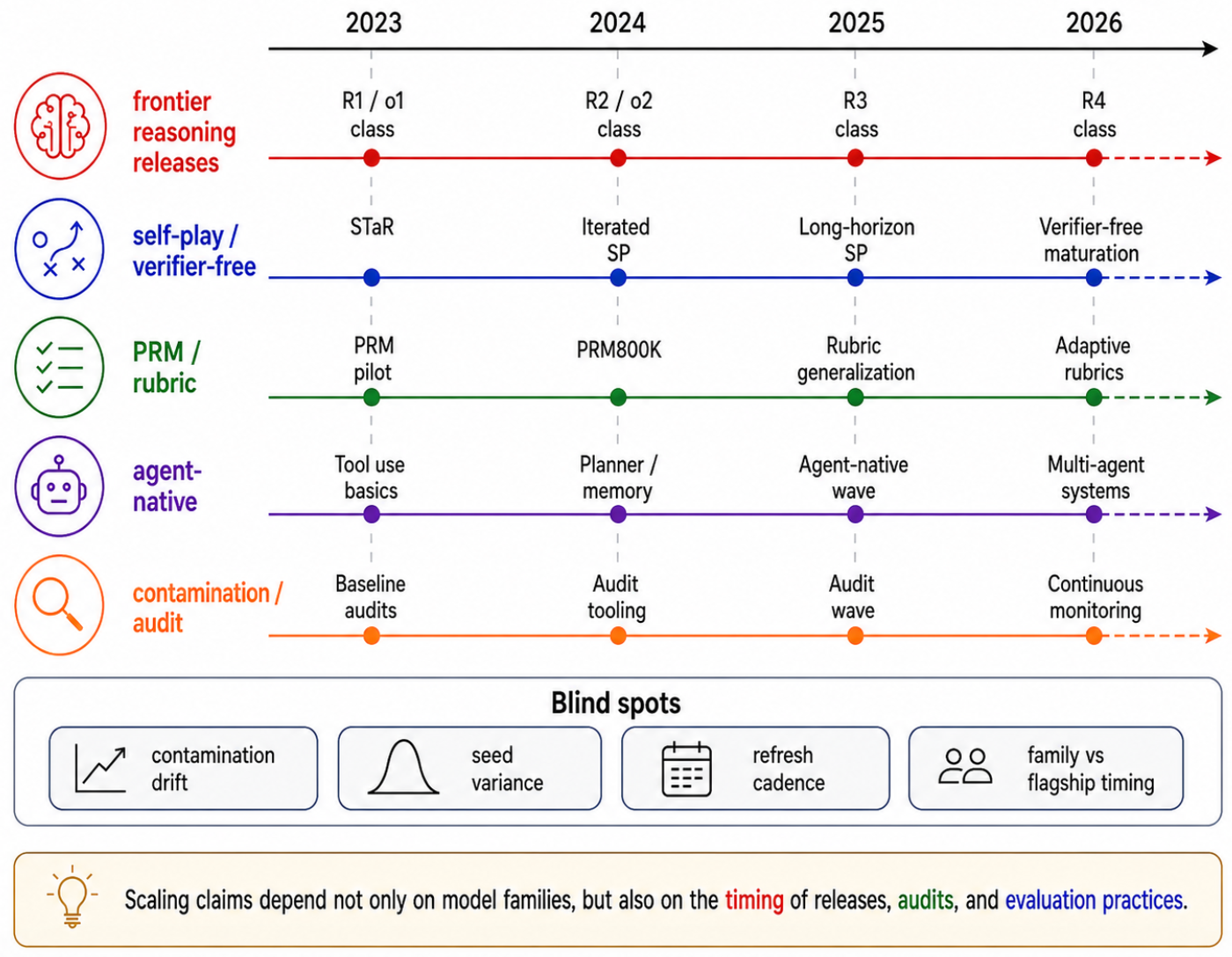}
    \caption{\textbf{Release timeline as a scaling object.}
    A release tick is a versioned data artifact, not a neutral timestamp. 
    Without prompt-pool, verifier-refresh, contamination, lineage, and evaluation-budget fields, score differences can collapse data-substrate changes, optimizer changes, and measurement changes.}
    \label{fig:release-timeline}
    \vspace{-0.25cm}
\end{figure}

Figure~\ref{fig:release-timeline} treats release time as part of the scaling object. 
Reasoning-model comparisons often mix model-family changes, data refreshes, verifier updates, benchmark audits, and evaluation-budget changes. 
A release date is therefore not a neutral timestamp: it can mark a different prompt pool, a new verifier, a changed inference budget, or a new contamination audit.

\subsection{Agent Trajectory Audit Fields}
\label{app:agent-audit}

Environmental reasoning data require a richer audit schema than prompt--answer datasets because the trainable object is a trajectory. 
A successful final transcript is often insufficient: it may hide failed tool calls, retries, state diffs, invalid actions, scaffold interventions, or terminal predicates. 
Table~\ref{tab:agent-audit-fields} lists fields that make agent trajectories replayable and auditable.

\begin{table*}[t]
\centering
\small
\setlength{\tabcolsep}{4pt}
\renewcommand{\arraystretch}{1.10}
\begin{tabularx}{\textwidth}{@{}
>{\raggedright\arraybackslash}p{0.16\textwidth}
>{\raggedright\arraybackslash}p{0.30\textwidth}
>{\raggedright\arraybackslash}X@{}}
\toprule
\textbf{Audit field} & \textbf{What should be released} & \textbf{Why it matters} \\
\midrule
Task state 
& initial state, files, UI state, repository snapshot, environment version 
& determines whether the episode can be reset and replayed \\

Goal and constraints 
& user goal, hidden tests, policy constraints, allowed/disallowed actions 
& separates task success from scaffold compliance \\

Action schema 
& tool/API schema, action grammar, argument format, timeout and permission rules 
& defines the action space available to the model \\

Observations 
& tool outputs, browser/app observations, terminal logs, screenshots, error messages 
& exposes the feedback actually available during the episode \\

State diffs 
& file changes, patches, database changes, UI transitions, environment deltas 
& makes credit assignment and recovery behaviour inspectable \\

Failures and retries 
& invalid actions, failed calls, rejected patches, recovery attempts, backtracking 
& preserves the branches usually erased by cleaned SFT traces \\

Terminal predicate 
& success condition, unit tests, grader, judge, environment predicate 
& defines what counts as completion or reward \\

Scaffold metadata 
& planner, tool wrapper, memory, prompt template, agent loop, stopping rule 
& prevents scaffold changes from being mistaken for model changes \\

Budget and sampling 
& rollout count, token budget, time limit, temperature, pass@$k$ or selector 
& makes inference-time compute comparable across systems \\

Lineage and split 
& generator, teacher, verifier version, filtering rule, split, contamination audit 
& supports attribution across generations and releases \\
\bottomrule
\end{tabularx}
\caption{\textbf{Agent trajectory audit fields.}
Agent data should be released as replayable episodes rather than cleaned successful transcripts. The fields below expose the state, actions, feedback, scaffold, budget, and lineage needed to attribute gains in environmental reasoning tasks.}
\label{tab:agent-audit-fields}
\vspace{-0.35cm}
\end{table*}

The table also clarifies why environmental verification is not reducible to final success. 
If a release includes only the successful path, it removes the branches where tool misuse, recovery, exploration, and credit assignment become visible. 
For post-training, those branches may be the most informative part of the data.

\subsection{Source Placement Guide}
\label{app:source-placement}

The paper uses sources in different roles depending on what each source makes visible. 
Model reports are useful for documenting public recipes and optimizer scaffolds, but they often under-specify data lineage. 
Dataset releases are useful for identifying data objects and verifier contracts, but they may not isolate causal training effects. 
Benchmark and evaluation papers reveal failure modes and measurement surfaces, while scaling studies help separate ceiling movement from approach efficiency. 
Table~\ref{tab:source-placement} summarizes how sources are placed in the primer.

\begin{table*}[t]
\centering
\small
\setlength{\tabcolsep}{4pt}
\renewcommand{\arraystretch}{1.10}
\begin{tabularx}{\textwidth}{@{}
>{\raggedright\arraybackslash}p{0.16\textwidth}
>{\raggedright\arraybackslash}p{0.27\textwidth}
>{\raggedright\arraybackslash}p{0.25\textwidth}
>{\raggedright\arraybackslash}X@{}}
\toprule
\textbf{Source role} & \textbf{What it contributes} & \textbf{Typical examples} & \textbf{Placement rule} \\
\midrule
Model reports 
& public post-training scaffold, recipe description, inference budget 
& DeepSeek-R1, Kimi K1.5, Qwen3-style reports \citep{deepseekr12025,kimi2025,yang2025qwen3} 
& use for visible pipeline claims, not alone for data-causal attribution \\

Reasoning-data releases 
& prompt source, answer format, verifier, filter, split, contamination audit 
& DeepMath-103K, DAPO, OpenThoughts \citep{he2025deepmath103k,dapo2025,openthoughts2025} 
& use for data-object and construction claims \\

Process-supervision work 
& step labels, PRM training, rollout values, first-error localization 
& PRM800K, Math-Shepherd, OmegaPRM \citep{lightman2023verify,mathshepherd2023,omegaprm2024} 
& use for trace-quality and supervision-granularity claims \\

Agent / environment benchmarks 
& state, action, observation, terminal predicate, replay or task scaffold 
& SWE-Gym, AppWorld, OSWorld-style tasks \citep{swegym2024,appworld2024,osworld2024} 
& use for environmental verification and trajectory-audit claims \\

Verifier and judge studies 
& verifier failure modes, judge bias, reward hacking, false rejection 
& verifier robustness, master-key, spurious reward studies \citep{huang2025verifierrobustness,zhao2025masterrm,shao2025spuriousrewards} 
& use for correctness and audit-risk claims \\

Scaling studies 
& compute, data reuse, asymptote, approach efficiency, inference-time budget 
& RL scaling and post-training scaling analyses \citep{khatri2025scalerl,tan2025scalingrl} 
& use for scaling-attribution claims \\
\bottomrule
\end{tabularx}
\caption{\textbf{Source placement guide.}
Sources are placed according to the kind of evidence they expose. 
This prevents model reports, data releases, verifier audits, and scaling studies from being treated as interchangeable evidence for the same claim.}
\label{tab:source-placement}
\vspace{-0.35cm}
\end{table*}

\subsection{Use of AI Assistance}
\label{app:ai-use}

The authors used AI assistants, including ChatGPT, for language polishing, structural editing, and drafting brief disclosure and limitation text. AI assistants were not used to generate empirical results, create datasets, or determine source inclusion, attribution claims, or citations. All claims, citations, and final manuscript wording were reviewed and verified by the authors.

\end{document}